\documentclass[english]{article}

\usepackage{algorithmic}
\usepackage{algorithm}
\usepackage{amsfonts}
\usepackage{amsmath}
\usepackage{amssymb}
\usepackage{amsthm,wrapfig}
\usepackage{array}
\usepackage{arydshln}
\usepackage{bm}
\usepackage{cite}
\usepackage{color}
\usepackage{comment}
\usepackage{dsfont}
\usepackage{enumitem}
\usepackage{float}
\usepackage[T1]{fontenc}
\usepackage{geometry}
\usepackage{graphicx}
\usepackage{grffile}
\usepackage{hyperref}\hypersetup{colorlinks,linkcolor=blue,anchorcolor=blue,citecolor=blue}
\usepackage[latin9]{inputenc}
\usepackage{letltxmacro}
\usepackage{mathrsfs}
\usepackage{mathtools}
\usepackage{multirow}
\usepackage{nicefrac}
\usepackage{subcaption}
\usepackage{tablefootnote}
\usepackage{verbatim}

\usepackage{todonotes}

\usepackage{booktabs}

\allowdisplaybreaks

\newcommand{\bs}{\bm{s}}

\newcommand{\bx}{\bm{x}}

\newcommand{\bz}{\bm{z}}

\newcommand{\bA}{\bm{A}}
\newcommand{\bB}{\bm{B}}

\newcommand{\bL}{\bm{L}}

\newcommand{\bR}{\bm{R}}

\newcommand{\bW}{\bm{W}}
\newcommand{\bX}{\bm{X}}

\newcommand{\bZ}{\bm{Z}}

\newcommand{\cX}{\mathcal{X}}

\newcommand{\RR}{\mathbb{R}}

\newcommand{\one}{\bm{1}}

\allowdisplaybreaks
\makeatletter

\theoremstyle{plain} 

\theoremstyle{definition}
 
\theoremstyle{remark}

\definecolor{tian}{RGB}{0,150,0}
\definecolor{cm}{RGB}{250,0,200}
\definecolor{yc}{RGB}{255,0,0}
\definecolor{hd}{RGB}{0,180,200}

\definecolor{edits}{RGB}{250,100,100}

\newcommand{\methodname}{\texttt{Caprese}}
\newcommand{\griffin}{GRIFFIN}

\begin{document}
\title{Scalable LLM Reasoning Acceleration with\\Low-rank Distillation}

\author
 {
 	Harry Dong\thanks{Department of Electrical and Computer Engineering, Carnegie Mellon University, USA; Email: \texttt{harryd@andrew.cmu.edu}.} \\
	CMU 
	\and
    Bilge Acun\thanks{FAIR at Meta} \\
    Meta
    \and
 	Beidi Chen\footnotemark[1] \\
 	CMU
    \and
    Yuejie Chi\footnotemark[1] \footnotemark[2] \\
    CMU, Meta
 }

\date{\today}

\setcounter{tocdepth}{2}
\maketitle

\begin{abstract}

Due to long generations, large language model (LLM) math reasoning demands significant computational resources and time. While many existing efficient inference methods have been developed with excellent performance preservation on language tasks, they often severely degrade math performance. In this paper, we propose \methodname{}, a resource-efficient distillation method to recover lost capabilities from deploying efficient inference methods, focused primarily in feedforward blocks. With original weights unperturbed, roughly 1\% of additional parameters, and only 20K synthetic training samples, we are able to recover much if not all of the reasoning capabilities lost from efficient inference for thinking LLMs and without harm to language tasks for instruct LLMs. Moreover, \methodname{} slashes the number of active parameters ($\sim$2B cut for Gemma 2 9B and Llama 3.1 8B) and integrates cleanly into existing model layers to reduce latency (>16\% time-to-next-token reduction) while encouraging response brevity (up to 8.5\% fewer tokens).

\end{abstract}



\section{Introduction}
\label{sec:intro}

With the increasing capabilities of large language models (LLMs) \cite{vaswani2017attention}, outputs are also becoming increasingly sophisticated, typically involving multi-step reasoning such as in math problem solving. These tasks tend to demand long generation which drives up latency, making efficiency a dire issue. Fortunately, many sparsity-based efficient LLM inference algorithms have shown great promise, slashing expensive computational bottlenecks with little damage to the original performance on a variety of language-based tasks like reading comprehension and summarization. \textit{However, for reasoning tasks, many of these algorithms begin to break down, decimating performance, despite their robustness in language settings.} Thus, there is a need to design efficient inference algorithms that simultaneously maintain language and reasoning capabilities.

One of the main differences between reasoning and many language tasks is the generation length. Reasoning typically involves long generations from chain-of-thoughts (CoTs) \cite{wei2022chain}, which often far surpass the length of the input query. However, CoT performance falls apart with efficient algorithms that introduce approximation errors which build up over time. For instance, deploying CATS \cite{lee2024cats}, a sparse thresholding method, on Gemma 2 2B \cite{team2024gemma2} has little impact on language generation performance, but knocks GSM8K \cite{cobbe2021gsm8k} accuracy \textit{from 51.02\% to 34.42\%} (Table~\ref{tab:instruct_results_small}). Similarly, for thinking models: applying \griffin{} \cite{dong2024prompt}, an adaptive structured pruning method, on the first half of DeepSeek-R1-Distill-Qwen 1.5B \cite{guo2025deepseek} drops MATH-500 \cite{lightman2023lets} accuracy \textit{from 79.40\% to 42.00\%} (Table~\ref{tab:thinking_results}). Moreover, since generation is much more computationally demanding than prefill, there is a dire need to make long reasoning CoTs efficient, which is made more apparent with the rise of test-time scaling.
Thus, this poses two key inference challenges with math reasoning. 
First, even single token mistakes can sometimes drive the generation trajectory off course, leading to an incorrect response \cite{zhou2024sirius}. Second, the already computationally expensive LLM autoregressive decoding process is accentuated with generating CoTs. This problem is further exaggerated as CoTs scale in length in thinking models \cite{guo2025deepseek} and as the number of repeated generations scale to elicit higher quality answers \cite{brown2024large, wu2024empirical, snell2024scaling}. \textit{An ideal method should be efficient, performance preserving, and easy to integrate.}

\begin{figure}[t]
\begin{center}
\includegraphics[width=0.85\columnwidth]{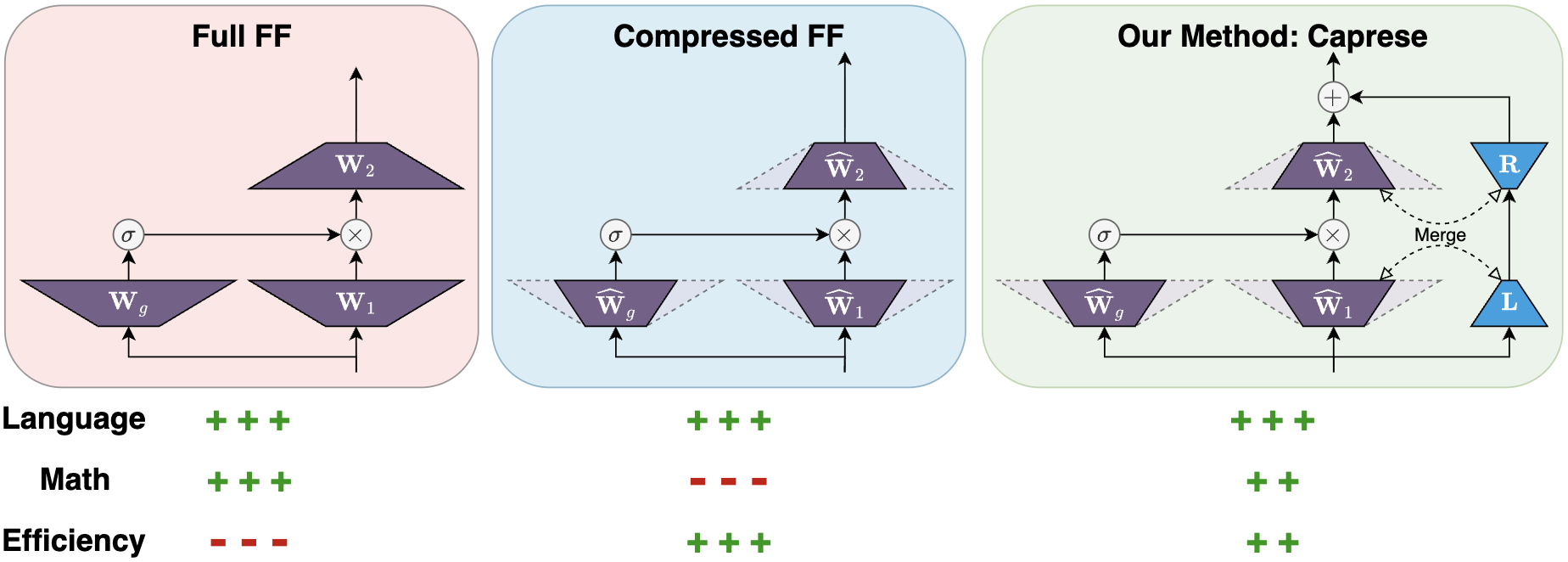}
\caption{A full FF block maximizes accuracy without any benefit to efficiency. Compressed FF algorithms can be very efficient by using subsets of the FF block but harm math performance. Our method, \methodname{}, uses a compressed FF algorithm and a small distilled low-rank linear layer, which can be merged with existing FF weights, for performative inference in language and math settings while being efficient. Layers are drawn as trapezoids to highlight the expansion of the intermediate feature size compared to the input size.}
\label{fig:method_comparison}
\end{center}
\vspace{-0.25in}
\end{figure}

\begin{figure}[h]
\begin{center}
\includegraphics[width=0.35\columnwidth]{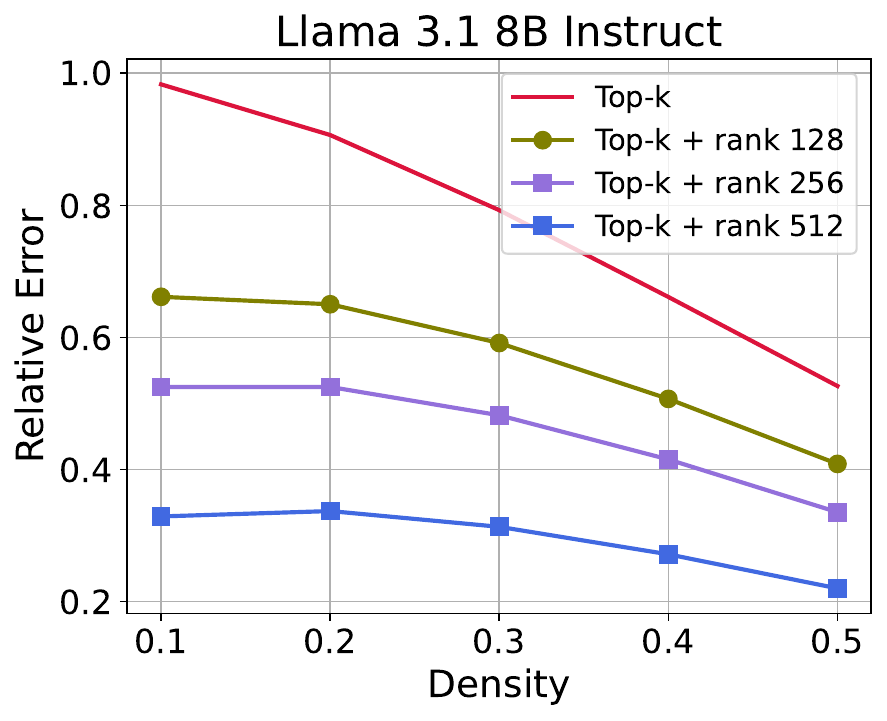}
\includegraphics[width=0.35\columnwidth]{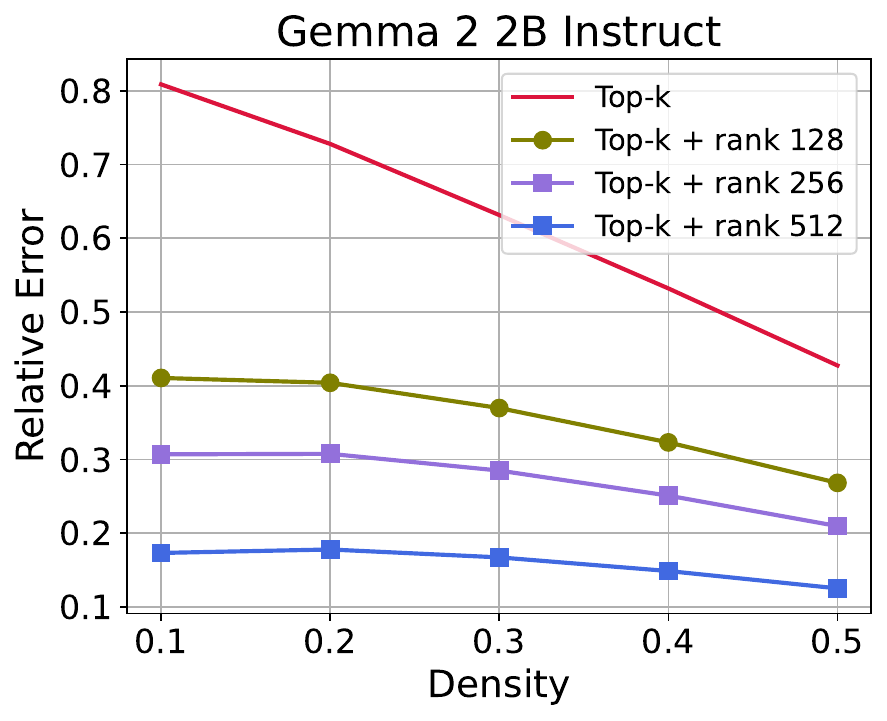}
\vspace{-0.1in}
\caption{Average relative FF output error of generated tokens with varying top-$k$ densities and low-rank approximations. The density is the fraction of non-zero intermediate FF features maintained by top-$k$. Samples consist of 16 random MATH generations by the original model. \textit{A relatively small low-rank approximation to the top-$k$ residual reduces error more effectively than increasing density.}}
\label{fig:topk_vs_error}
\end{center}
\vspace{-0.1in}
\end{figure}

Thankfully, low-rank structure in the feedforward (FF) output features can help make this possible. (We choose to focus on FF blocks since they contribute around 2/3 of an LLM's parameters and about 50\% of the generation latency.) Because of the success of works that exploit contextual sparsity in FF blocks on language tasks for efficiency, like CATS and \griffin{}, we peek into the residuals of FF sparsity-based efficient methods. Using an oracle top-$k$ filter on FF nonlinearity output magnitudes, we observe huge reductions in error with a low-rank approximation to the FF output residuals (Figure~\ref{fig:topk_vs_error}). Since FF intermediate feature sizes can be on the order of $10^5$, adding 256 for a low-rank approximation is comparatively tiny. \textit{This observation motivates us to estimate the residual from FF compression methods with low-rank layers.}

We introduce \textbf{\methodname{}} (\textbf{CAP}ability \textbf{RE}covery with \textbf{S}calable \textbf{E}fficiency) to learn FF residuals from a compressed FF algorithm using small low-rank linear layers for improved performance (Figure~\ref{fig:method_comparison}). While \methodname{} works for any task, we focus on reasoning (e.g., math) in this paper, as current efficient compression methods can obliterate an LLM's reasoning capabilities. Through distillation of math knowledge into these low-rank layers, \methodname{} is able to overcome the aforementioned challenges of reasoning and makes significant progress towards an ideal efficient method:
\begin{enumerate}
    \item \textbf{Performance Enhancement:} Demonstrated across a variety of instruct and thinking models, \methodname{} \textit{recovers much if not all of the reasoning performance lost from deploying a compressed FF algorithm without harming language tasks}. Also, \methodname{} maintains its performance benefit when applying different techniques of test-time scaling. 
    \item \textbf{Efficiency:} Due to its parallelizability with existing FF layers, \methodname{} adds negligible overhead, preserving latency reductions of the underlying method. 
    \item \textbf{Low Budget:} The distillation process of \methodname{} is cheap since we are able to see significant gains in performance by training on only 20K synthetic math samples. Moreover, with a low-rank layer size of only 256, this equates to adding roughly $0.8\%$ additional parameters into Llama 3.1 8B and Gemma 2 9B, dwarfed by savings in active parameters ($\sim$2B cut for both models with \methodname{} (CATS)).
\end{enumerate}

Our extensive experiments on \methodname{} indicate strong performance on various tasks and models. For example, DeepSeek-R1-Distill-Qwen 7B with a sparse method drops AMC 2023 accuracy from 75.00\% to 62.50\%, but \methodname{} is able to bring it to 78.13\%, beyond the original accuracy. Furthermore, scaling generation quantity with \methodname{} on Llama 3.2 3B Instruct improves Pass@$100$ by 7.0\% from the original model while using 15.8\% fewer active parameters. We also show \methodname{} reduces latency by 16\% using the Qwen 2.5 14B architecture.

\paragraph{Paper Organization.} Section~\ref{sec:background} covers related works and describes \griffin{} and CATS, two efficient sparse FF compression algorithms that serve as baselines. Section~\ref{sec:method} details \methodname{}'s architecture and distillation procedure. Then, we showcase the strong performance of \methodname{} in Section~\ref{sec:experiments}: scaling generation length with thinking LLMs (Section~\ref{sec:thinking_experiments}), scaling generation outputs (Section~\ref{sec:coverage_experiments}), and instruct LLM inference (Section~\ref{sec:instruct_experiments}). Finally, we quantify efficiency improvements in Section~\ref{sec:efficiency}.

\section{Background}
\label{sec:background}

Here, we provide an overview of the efficient LLM inference literature that \methodname{} builds upon with in-depth descriptions of \griffin{} and CATS which we use as baselines.

\subsection{Related Works}

\paragraph{Efficient Inference for FF Blocks.}
To improve the efficiency of FF blocks in LLMs, various methods leverage existing sparse structures in FF features \cite{geva2020transformer, dettmers2022llm, li2022lazy, dong2023towards, liu2023deja, sadhukhan2026stem}. Pruning sets a parameter subset to zero to enforce static sparsity \cite{lecun1989optimal, sun2023simple, frantar2023massive, ma2023llm}. Mixtures of experts (MoEs) adaptively select predefined parameter subsets in FF blocks, though they tend to require significant fine-tuning or training from scratch \cite{jacobs1991adaptive, zhang2021moefication, dai2024deepseekmoe, liu2023deja}. Similar to MoEs, another line of work aims at exploiting contextual sparsity to dynamically select FF neurons for each input without training. \griffin{} \cite{dong2024prompt} is a calibration-free method that uses FF activations from the prefill phase to adaptively prune FF neurons for generation. CATS \cite{lee2024cats} uses hard thresholding to skip computation in parts of the FF block with a custom kernel. Though not constrained to FF blocks, quantization is another popular method for compression \cite{dettmers2023qlora,dettmers2022gpt3, dong2024towards,lin2024awq}.

\paragraph{Shorter CoTs.}
Several works aim to improve reasoning efficiency by directly reducing the number of generated tokens \cite{renze2024benefits, nayab2024concise,arora2025training, aytes2025sketch, xu2025chain, qu2025optimizing}. Since we aim to reduce the per-token latency, this line of research to encourage brevity is orthogonal to ours and could be used alongside our method.

\paragraph{Test-time Scaling.}
To enhance LLM performance, the priority has historically been to scale training with more data and bigger models \cite{kaplan2020scaling, hoffmann2022training}. More recently, there is an increasing effort towards test-time scaling to boost performance on difficult tasks like math. Two ways of test-time scaling that have seen great success. First, for a single prompt, multiple responses can be sampled from the LLM \cite{brown2024large, snell2024scaling, wu2024empirical, manvi2024adaptive, sun2024fast, dong2025generalized}. This increases the probability that a desired response lies in the pool of responses. While this method of scaling is highly parallelizable, it also relies on a verifier model to select the best one. Second, CoTs can be lengthened for each response \cite{guo2025deepseek, muennighoff2025s1,yang2025qwen3}. By extrapolating the success of CoTs to extreme lengths (e.g., DeepSeek-R1 generates up to 32K tokens per response), the accuracy of the final answer improves significantly, but this method has low parallelizability due to the autoregressive nature of generation.


\subsection{Training-free \& Adaptively Sparse FF Compression Methods}
\label{sec:ff_algorithms}

We now briefly describe the inner workings of \griffin{} and CATS. Let $\bX \in \RR^{S \times D}$ be the input into the FF block during the prefill phase with sequence length $S$ and feature size $D$. Define the FF block as $\text{FF}(\bX) = \text{FF}_2(\text{FF}_1(\bX))$ such that
\begin{align}
    \bZ = \text{FF}_1(\bX) &= \sigma(\bX \bW_g) \odot \bX \bW_1, \\
    \text{FF}_2(\bZ) &= \bZ \bW_2,
\end{align}
where $\bW_g, \bW_1 \in \RR^{D \times D_{\text{FF}}}$, $\bW_2 \in \RR^{D_{\text{FF}} \times D}$, $\sigma$ is a nonlinear function, $\odot$ is an element-wise multiplication operator, and $D_{\text{FF}}$ is the FF intermediate feature size. Typically, $D_{\text{FF}} \gg D$. Although recent LLMs use this architecture, for LLMs without gated functions (e.g., OPT \cite{zhang2022opt}), $\bX \bW_1$ can be removed. Bias terms are omitted for brevity.

\paragraph{\griffin{}.}
\griffin{} \cite{dong2024prompt} adaptively prunes columns and rows in the FF block weights, using FF activation statistics from the prefill phase, namely the flocking patterns. \griffin{} calculates $[\overline{\bZ}]_i = |[\bZ]_i| / \|[\bZ]_i\|_2$ for each token index $i$, followed by an aggregation across the FF feature axis: $[\bs]_j = \|[\overline{\bZ}]_{\cdot, j}\|_2$. The result $\bs \in \RR^{D_\text{FF}}$ gives a metric to perform top-$k$ selection across corresponding columns of $\bW_g$ and $\bW_1$, and rows of $\bW_2$ to produce $\widehat{\bW}_g, \widehat{\bW}_1 \in \RR^{D \times k}$, and $\widehat{\bW}_2 \in \RR^{k \times D}$. Then, for the generation phase, the following FF block is used for input $\bx \in\RR^{D}$:
\begin{align}
    \widehat{\bz} = \widehat{\text{FF}}_1(\bx) &= \sigma(\bx \widehat{\bW}_g) \odot  \bx \widehat{\bW}_1, \\
    \widehat{\text{FF}}_2(\widehat{\bz}) &=   \widehat{\bz} \widehat{\bW}_2.
\end{align}
The compressed FF blocks are fixed throughout generation but are dynamic across prompts.

\paragraph{CATS.}
CATS \cite{lee2024cats} uses hard thresholding to skip computation in part of the FF block. Letting $T_\tau$ be the hard thresholding function with threshold $\tau$, CATS computes 
\begin{align}
    \widehat{\bz} = \widehat{\text{FF}}_1(\bx) &= T_\tau(\sigma(\bx \bW_g)) \odot \bx \bW_1, \\
    \widehat{\text{FF}}_2(\widehat{\bz}) &= \widehat{\bz} \bW_2.
\end{align}
The weights $\bW_1$ and $\bW_2$ should be sparsified into $\widehat{\bW}_1$ and $\widehat{\bW}_2$, respectively, based on the non-zero entries of $T_\tau(\sigma(\bx \bW_g))$ for latency improvement, which can vary from token to token and require a custom kernel for wall clock speed-up. The parameter $\tau$ is calibrated to be a desired percentile on a dataset. In this paper, to avoid calibration, we set $\tau$ based on prefill features and only threshold during the generation phase, analogous to \griffin{}.

\section{Method: \methodname{}}
\label{sec:method}

Motivated by the low-rank structure of residuals in Figure~\ref{fig:topk_vs_error}, we introduce \methodname{} which distills approximation errors in embeddings into low-rank linear layers in FF blocks. See Figure~\ref{fig:method_comparison} for an illustration of our method.

\subsection{Layer-wise Distillation}

Inference efficiency algorithms often introduce feature approximation errors in favor of faster generation, which we mitigate with distillation. Let the efficient and approximate FF block be $\widehat{\text{FF}}(\bx) = \widehat{\text{FF}}_2(\widehat{\text{FF}}_1(\bx))$. In our design of \methodname{}, we do not constrain $\widehat{\text{FF}}$ to be any specific method or architecture. For instance, $\widehat{\text{FF}}$ could be an FF block with \griffin{} \cite{dong2024prompt}, CATS \cite{lee2024cats}, or quantized weights. Then, the error is 
\begin{align*}
    \| \text{FF}(\bx) - \widehat{\text{FF}} (\bx) \|_2^2 .
\end{align*}
We choose to reduce this residual with a low-rank linear layer, meaning we want to solve
\begin{align}\label{eq:layerwise_obj}
    \min_{\bL, \bR} \frac{1}{|\cX|}\sum_{\bx \in \cX} \| \text{FF}(\bx) - \widehat{\text{FF}} (\bx) - \bx \bL \bR \|_2^2
\end{align}
for input set $\cX$, $\bL \in \RR^{D \times r}$, and $\bR \in \RR^{r \times D}$ where $r \ll D_{\text{FF}}$. The optimal solution can be computed analytically since this is a reduced rank regression problem, but the size of $|\cX|$ and $D$ may make it prohibitively expensive. Therefore, we opt to learn $\bL$ and $\bR$ independently for every FF block (i.e., previous FF blocks are assumed to be from the original model), allowing for parallel layer-wise training. This takes inspiration from LESS \cite{dong2024get} which uses layer-wise training on attention residuals for key-value cache compression. Each $\bR$ is initialized as a zero matrix since the efficient approximation is assumed to be of good quality, and original model weights are frozen. We also distill end-to-end (E2E) to further improve the performance.


\subsection{End-to-end Distillation}

Using the learned low-rank layers as an initialization, we put them all together to distill the final model embedding before the linear head into the efficient model, again using MSE: 
\begin{align}
    \min_{(\bL_i, \bR_i)_{i = 1, \dots, L}} \frac{1}{|\cX|}\sum_{\bx \in \cX} \| \text{M}(\bx) - \text{M}_{\text{student}}(\bx) \|_2^2
\end{align}
where $\text{M}$ and $\text{M}_{\text{student}}$ are the original LLM and efficient LLM with distillation layers, respectively, excluding the final linear head. $L$ is the number of layers in the model. All original weights are frozen, so the only tunable parameters are the $\bL$ and $\bR$ of each FF block.

\subsection{Parallel Inference Computation}
\label{sec:parallel_compute}

The computation of $\bx \bL \bR$ can be done in parallel with the original FF operations. In fact, $\bL$ and $\bR$ can be concatenated with the up and down projection matrices, respectively. 

In other words, the \methodname{} FF block is $\widehat{\text{FF}}^+(\bx) = \widehat{\text{FF}}^+_2(\widehat{\text{FF}}^+_1(\bx))$, such that
\begin{align}
    \widehat{\bz}^+ = \widehat{\text{FF}}^+_1(\bx) &= \begin{bmatrix}\sigma(\bx \widehat{\bW}_g) & \one_r \end{bmatrix} \odot \bx \begin{bmatrix} \widehat{\bW}_1 &\bL \end{bmatrix}, \\
    \widehat{\text{FF}}^+_2(\widehat{\bz}^+) &=   \widehat{\bz}^+ \begin{bmatrix} \widehat{\bW}_2^\top &\bR^\top \end{bmatrix}^\top,
\end{align}
where $\one_r$ is a one-vector with length $r$. In practice, to save memory and time, we do not materialize $\one_r$ but directly assign the product with $\sigma(\bx \widehat{\bW}_g)$ to corresponding entries of $\bx \widehat{\bW}^+_1$.
Recall that the prefill stage still just uses the original model.

\subsection{Training Details}

We set the inner dimension of our low-rank layer to $r=256$ (to see the effect of different $r$, see Appendix~\ref{app:rank_v_sparsity}). In comparison to the large inner dimension of FF layers (e.g., $D_{\text{FF}} = 14336$ for Llama 3.1 8B and Gemma 2 9B), our choice of $r$ is relatively miniscule, adding only $\sim 1\%$ new parameters for all tested models (Appendix~\ref{app:parameters}). We use a 20K subset of a common synthetic math training set for training. For a fair comparison, the same subset is used for both layer-wise and E2E distillation for every model. Each training sample is prepended with a CoT instruction: ``Please reason step by step.'' At test time, the actual instructions may be vastly different. Layer-wise and E2E training consists of 20 epochs and 3 epochs, respectively. Training and inference are done in BF16. Hyperparameters are listed in Appendix~\ref{app:hyperparameters}.

\subsection{Comparison with LoRA}
\label{sec:lora}
Our method derives inspiration from and shares a slight connection with low-rank adaptation (LoRA), a widely used method for efficient fine tuning with the addition of low-rank parameters \cite{hu2022lora}, but remains distinct since they target different inefficiencies. 
To illustrate, recall a sparse algorithm constructs $\widehat{\bW}_1 \in \RR^{D \times k}$ by selecting columns from $\bW_1$. Alternatively, construct $\widetilde{\bW}_1 \in \RR^{D \times D_{\text{FF}}}$ by setting unwanted columns to zero, so nonzero columns in $\widetilde{\bW}_1$ match columns in $\widehat{\bW}_1$ and vice versa. Doing the same to columns of $\bW_g$ and rows of $\bW_2$, LoRA learns $\bA_1, \bA_g, \bB_2^\top \in \RR^{D \times r}$ and $\bB_1, \bB_g, \bA_2^\top \in \RR^{r \times D_{\text{FF}}}$ to form a new FF block with LoRA weights:
\begin{align}
    \widetilde{\text{FF}}(\bx) = \left(\sigma \big(\bx (\widetilde{\bW}_g + \bA_g \bB_g )\big) \odot \big(\bx (\widetilde{\bW}_1 + \bA_1 \bB_1) \big) \right) (\widetilde{\bW}_2  + \bA_2 \bB_2).
\end{align}
From this, we see \textit{LoRA is not designed to be applied to sparse efficient inference algorithms}. First, this is because in its compressed form, LoRA requires two sequential operations per linear layer. In contrast, \methodname{} parameters can be appended to existing ones, so additional computation is all parallel to the original operations (Section~\ref{sec:parallel_compute}). The real latency impact of this difference is quantified later in Table~\ref{tab:griffin_efficiency}. \methodname{}'s architecture also allows efficient training, but that is merely a byproduct of our design, not its purpose. Second, for a method like GRIFFIN which reduces the intermediate FF feature size from $D_{\text{FF}}$ to $k$, adding LoRA will negate this benefit while \methodname{} will preserve the reduced feature size. Due to this difference, we de-prioritize comparisons with LoRA in the main text, but include comparisons with LoRA in Appendix~\ref{app:lora} for interested readers.

\begin{table}[t]
\caption{Thinking models' 0-shot accuracies on reasoning tasks. Sparsity is set at 50\%, and $r = 256$. AIME 2024, AMC 2023, and BRUMO 2025 columns also include the sample standard deviation across 4 runs. Further improvements with reselection are shown in Table~\ref{tab:griffin_reselection}.}
\vspace{-0.15in}
\begin{center}
\begin{tabular}{lccccc}
\toprule
Model  & MATH-500 & AIME 2024 & AMC 2023 & BRUMO 2025 & GPQA \\
\midrule

\textit{DS-Qwen 1.5B}                    & 79.40 & 30.00 {\tiny $\pm$ 2.72}	& 60.83 {\tiny $\pm$ 7.47} & 25.83 {\tiny $\pm$ 3.19} & 18.69 \\
\hdashline
\griffin{}               & 42.00 & 2.50 {\tiny $\pm$ 1.67} & 16.88 {\tiny $\pm$ 1.25} & 1.67 {\tiny $\pm$ 1.92} & 11.62 \\
Layer-wise \methodname{}     & 47.20 & 2.50 {\tiny $\pm$ 1.67} & 28.13 {\tiny $\pm$ 2.39} & 0.00 {\tiny $\pm$ 0.00} & 13.13 \\
E2E \methodname{}     & \textbf{60.40} & \textbf{6.67} {\tiny $\pm$ 2.72} & \textbf{34.38} {\tiny $\pm$ 6.57} & \textbf{4.17} {\tiny $\pm$ 1.67}  & \textbf{16.67} \\
\hdashline
CATS                            & 72.00 & 11.67 {\tiny $\pm$ 6.94}	& 37.50 {\tiny $\pm$ 3.54} & 10.83 {\tiny $\pm$ 4.19} & 11.62\\
Layer-wise \methodname{}                  & 73.80 & 15.83 {\tiny $\pm$ 3.19} & \textbf{48.13} {\tiny $\pm$ 3.75} & 13.33 {\tiny $\pm$ 2.72} & 16.67 \\
E2E \methodname{}                  & \textbf{74.80} & \textbf{20.83} {\tiny $\pm$ 1.67}	& 47.50 {\tiny $\pm$ 3.54} & \textbf{20.00} {\tiny $\pm$ 2.72} & \textbf{22.22} \\
\midrule

\textit{DS-Qwen 7B}                    & 90.20 & 47.50 {\tiny $\pm$ 7.87}	& 75.00 {\tiny $\pm$ 4.08} & 45.83 {\tiny $\pm$ 4.19} & 38.38\\
\hdashline
\griffin{}               & 80.60 & 21.67 {\tiny $\pm$ 4.30}	& 62.50 {\tiny $\pm$ 8.90} & 25.00 {\tiny $\pm$ 8.39} & 21.72 \\
Layer-wise \methodname{}     & 84.80 & 29.17 {\tiny $\pm$ 3.19} & 64.38 {\tiny $\pm$ 7.18} & 27.50 {\tiny $\pm$ 4.19} & \textbf{27.78} \\
E2E \methodname{}     & \textbf{85.40} & \textbf{30.00} {\tiny $\pm$ 6.09}	& \textbf{71.88} {\tiny $\pm$ 5.54} & \textbf{29.17} {\tiny $\pm$ 5.69} & 23.74 \\
\hdashline
CATS                            & 89.20 & 34.17 {\tiny $\pm$ 7.88}	& 62.50  {\tiny $\pm$ 2.04} & 37.50 {\tiny $\pm$ 7.39} & 32.83 \\
Layer-wise \methodname{}                  & 87.00	& \textbf{35.00} {\tiny $\pm$ 4.30}	& 70.00 {\tiny $\pm$ 10.21} & 39.17 {\tiny $\pm$ 3.19} & \textbf{38.38} \\
E2E \methodname{}                  & \textbf{90.00} & 33.33 {\tiny $\pm$ 4.71} & \textbf{78.13} {\tiny $\pm$ 5.15} & \textbf{47.50} {\tiny $\pm$ 5.00} & 35.86 \\
\midrule

\textit{DS-Qwen 14B}                    & 92.80 & 63.33 {\tiny $\pm$ 2.72}	& 90.63 {\tiny $\pm$ 2.39} & 60.83 {\tiny $\pm$ 4.19} & 52.53 \\
\hdashline
\griffin{}               & 89.80	& 32.50 {\tiny $\pm$ 7.39}	& 80.63 {\tiny $\pm$ 3.75} & 33.33 {\tiny $\pm$ 3.85} & 41.92 \\
Layer-wise \methodname{}     & \textbf{90.80}	& \textbf{41.67} {\tiny $\pm$ 5.77} & \textbf{86.88} {\tiny $\pm$ 4.27} & \textbf{43.33} {\tiny $\pm$ 4.71} & \textbf{55.05} \\
E2E \methodname{}     & 89.20	& 39.17 {\tiny $\pm$ 9.18}	& 84.38 {\tiny $\pm$ 6.57} & 40.00 {\tiny $\pm$ 2.72} & 43.43 \\
\hdashline
CATS                            & \textbf{92.80}	& 58.34 {\tiny $\pm$ 1.92}	& \textbf{88.75} {\tiny $\pm$ 1.44} & \textbf{55.00} {\tiny $\pm$ 3.33} & 44.95 \\
Layer-wise \methodname{}                  & 91.00	& \textbf{61.67} {\tiny $\pm$ 5.77}	& \textbf{88.75} {\tiny $\pm$ 1.44} & 53.00 {\tiny $\pm$ 7.20} & 50.51 \\
E2E \methodname{}                  & 92.00	& 58.34 {\tiny $\pm$ 9.62} & 87.50 {\tiny $\pm$ 2.04} & 49.17 {\tiny $\pm$ 4.19} & \textbf{51.01} \\

\bottomrule
\end{tabular}
\end{center}
\label{tab:thinking_results}
\vspace{-0.2in}
\end{table}

\section{Experiments}
\label{sec:experiments}
We showcase the effectiveness of \methodname{} at recovering much, if not all, of the reasoning performance lost from efficient inference algorithms without sacrificing efficiency or performance on language tasks. Beginning with sequence length scaling in Section~\ref{sec:thinking_experiments}, \methodname{} is able to significantly or completely recover lost reasoning performance from existing efficient algorithms in thinking LLMs. Next, we demonstrate \methodname{}'s benefit when scaling the number of generations, another axis of tes-time scaling, in Section~\ref{sec:coverage_experiments}. Then in Section~\ref{sec:instruct_experiments}, we investigate typical instruct LLM settings without test-time scaling where we again observe better math performance without losing their generalizability. Finally, we highlight \methodname{}'s latency and length improvements in Section~\ref{sec:efficiency}. Although we focus on coupling \methodname{} with sparse FF methods, we show our method's merit carries over to quantization in Appendix~\ref{app:quantization}, broadening the potential applications of \methodname{}.
When ambiguous, we denote \methodname{} (CATS) to be our method with CATS as the underlying compression method and similarly for \griffin{}. Otherwise, we use ``\methodname{}'' for brevity. 
Unless specified, FF intermediate feature sparsity is set at 50\%. \textit{For a more meaningful baseline, we only apply \griffin{} to the first half of models as we observe steeper drops in math performance if used for all layers. CATS is applied to all layers.}

\subsection{Scaling Length: Thinking Models}
\label{sec:thinking_experiments}

Thinking models scale inference by augmenting the CoT length before giving a final answer. Since this entails long generation lengths, it is critical that error accumulation across tokens be minimized. We test on DeepSeek-R1-Distill-Qwen (which we abbreviate to DS-Qwen) models \cite{yang2024qwen2,guo2025deepseek} using Open R1 \cite{openr1} prompt templates and configurations (temperature and top-$p$ set to 0.6 and 0.95, respectively). We test 0-shot performance on math and science tasks MATH-500 \cite{hendrycks2021measuring, lightman2023lets}, AIME 2024 \cite{aime}, AMC 2023 \cite{amc23}, BRUMO 2025 \cite{brumo25}, and GPQA \cite{rein2024gpqa}, for a max generation length of 32768. Due to the small size of AIME 2024, AMC 2023, and BRUMO 2025, we evaluate them 4 times, reporting the average accuracy and sample standard deviation. We observe none of the methods have any significant effect on standard deviation in Table~\ref{tab:variance}.

From Table~\ref{tab:thinking_results}, \methodname{} is the most performative in most cases. DS-Qwen 1.5B and 7B see the greatest benefit with E2E \methodname{} usually achieving the highest accuracy, sometimes even exceeding the full model's performance. DS-Qwen 7B with CATS brings AMC 2023 accuracy down to 62.50\% from 75.00\%, but \methodname{} lifts performance above the original model's to 78.13\%, and similarly with DS-Qwen 7B with CATS and BRUMO 2025. All methods are more robust as model size increases. 
AIME 2024 and BRUMO 2025 challenge CATS and \griffin{} with severe degradation and partial recovery with \methodname{}. \textit{In the next section, we show a simple way to push this recovery even further with reselection.}

\begin{table}[h]
\caption{Average thinking model performance standard deviation on AIME 2024, AIME 2023, and BRUMO 2025 from Table~\ref{tab:thinking_results}. There is no clear consistent relationship between \methodname{} and accuracy variance.}
\vspace{-0.15in}
\begin{center}
\begin{tabular}{lcccc}
\toprule
 & DS-Qwen 1.5B & DS-Qwen 7B & DS-Qwen 14B & Mean \\
\midrule
Full Model & 4.46 & 5.38 & 3.10 & 4.31 \\
\hdashline
\griffin{} & 1.61 & 7.20 & 5.00 & 4.60 \\
Layer-wise \methodname{} & 1.35 & 4.85 & 4.92 & 3.71 \\
E2E \methodname{}& 3.65 & 5.77 & 6.16 & 5.19 \\
\hdashline
CATS & 4.89 & 5.77 & 2.23 & 4.30 \\
Layer-wise \methodname{} & 3.22 & 5.90 & 4.80 & 3.97 \\
E2E \methodname{} & 2.64 & 4.95 & 5.28 & 4.29 \\
\bottomrule
\end{tabular}
\end{center}
\label{tab:variance}
\vspace{-0.15in}
\end{table}

\subsubsection{Enhanced Performance with Reselection}

We can push the performance of \methodname{} with neuron reselection. For a sample, \griffin{} and CATS calculate metrics ($\bs$ and $\tau$) to determine subsets of the FF block to use, but these metrics are fixed during generation. Updating them mid-generation can benefit downstream performance.

For \griffin{}, updating the metric $\bs$ entails integrating the FF feature statistics of generated tokens into $\bs$. While this can be done by re-running prefill on all tokens, there is a more efficient way by passing in the generated tokens following the last reselection through the full model. As these tokens propagate through each layer, we find the selection metric for the generated tokens $\bs_G$ and update corresponding KV pairs.
This is like verification in speculative decoding \cite{chen2023accelerating,leviathan2023fast}. As $\bs$ and $\bs_G$ are $\ell_2$-norms along the token axis, we define the updated metric as $\sqrt{(\bs \odot \bs) + (\bs_G \odot \bs_G)}$ and use that to reselect different subsets of the FF block to use (as described in Section~\ref{sec:background}). \textit{This updates the pruned layers yet avoids prefill for all tokens.} Table~\ref{tab:griffin_reselection} shows a clear benefit of reselection (even if infrequent compared to speculative decoding) in \methodname{} by pushing the performance much closer to the full model's as we decrease steps between reselection rounds. Although \griffin{} is generally more harmful to accuracy than CATS due to its structured pruning, with reselection, \methodname{} (\griffin{}) can exceed the performance of \methodname{} (CATS) and reach the full model performance.

\begin{table}[t]
\caption{E2E \methodname{} AMC 2023 accuracies and standard deviations when recalculating \griffin{} pruning metrics and reselecting pruned neurons every $\rho$ decode steps. No reselection and the full model are special cases where $\rho = \infty$ and $\rho = 0$, respectively.}
\begin{center}
\begin{tabular}{lccccc}
\toprule
Model & No Reselect & $\rho = 1024$ & $\rho=256$ & $\rho=64$ & Full \\
\midrule
DS-Qwen 1.5B  & 34.38 & 41.87 & 45.00 & \textbf{58.75} & 60.83 \\
DS-Qwen 7B    & 71.88 & 69.25 & 73.13 & \textbf{73.75} & 75.00 \\
DS-Qwen 14B   & 84.38 & 88.13 & 88.75 & \textbf{91.88} & 90.63 \\
\bottomrule
\end{tabular}
\end{center}
\vspace{-0.18in}
\label{tab:griffin_reselection}
\end{table}

\begin{figure}[t]
\begin{center}
\includegraphics[width=0.3\columnwidth]{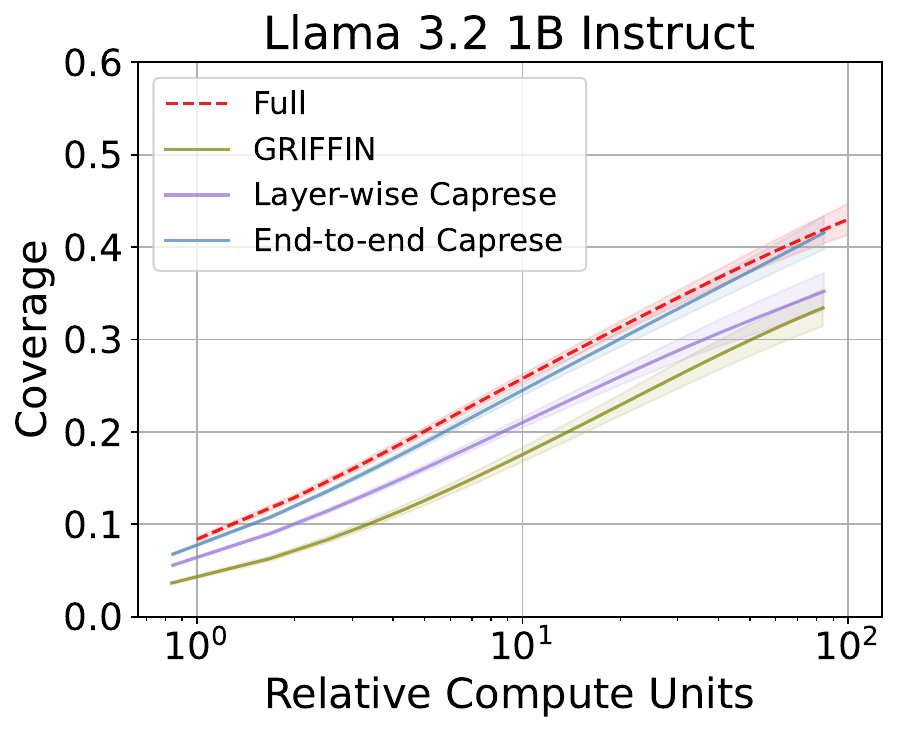}
\includegraphics[width=0.3\columnwidth]{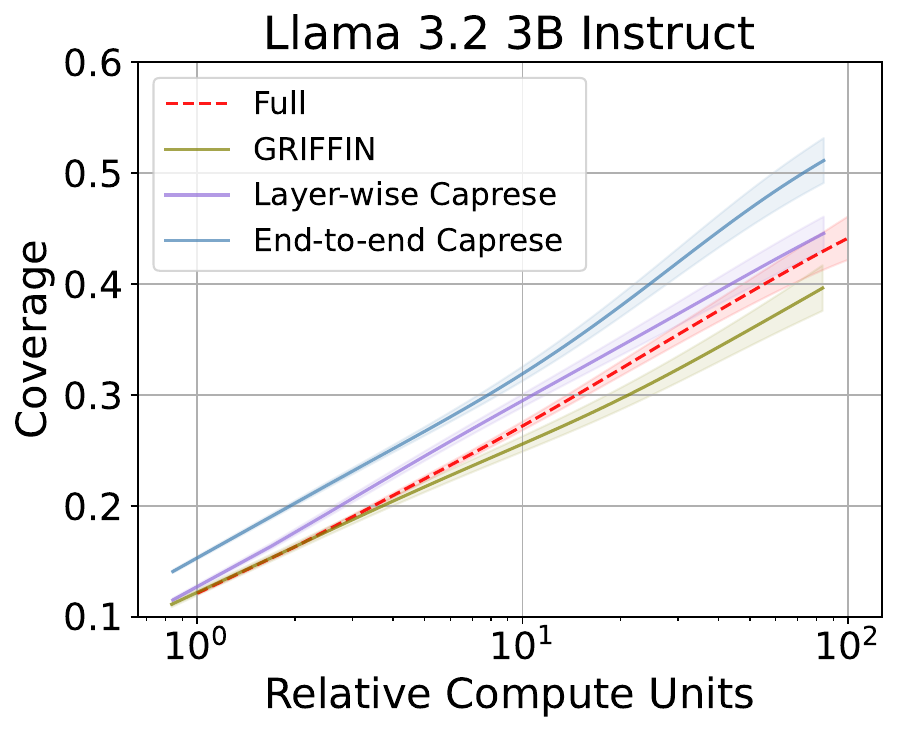}
\includegraphics[width=0.3\columnwidth]{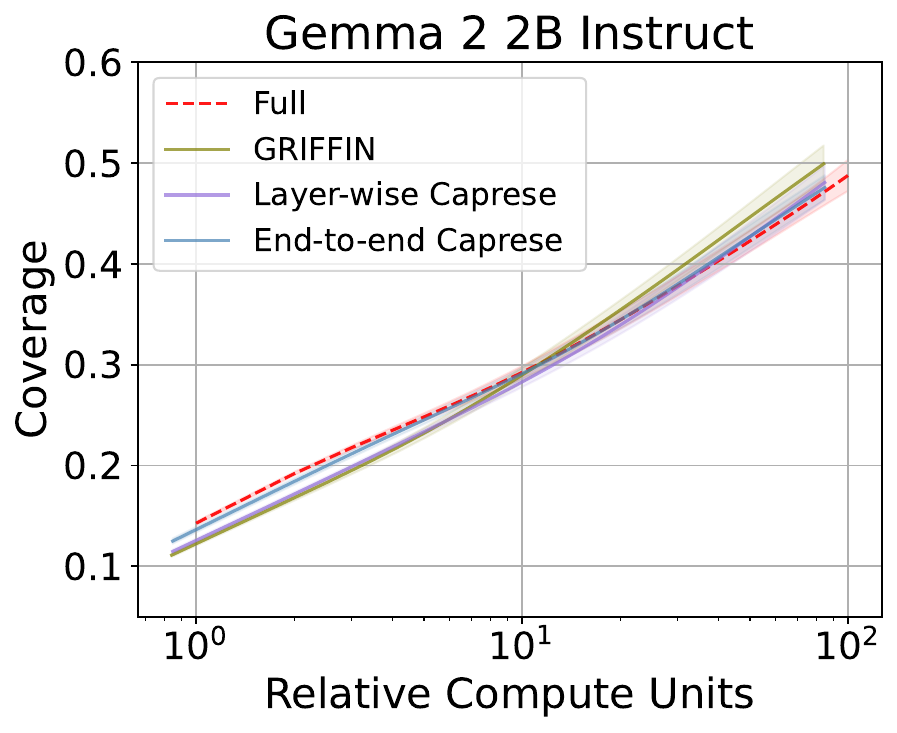}
\includegraphics[width=0.3\columnwidth]{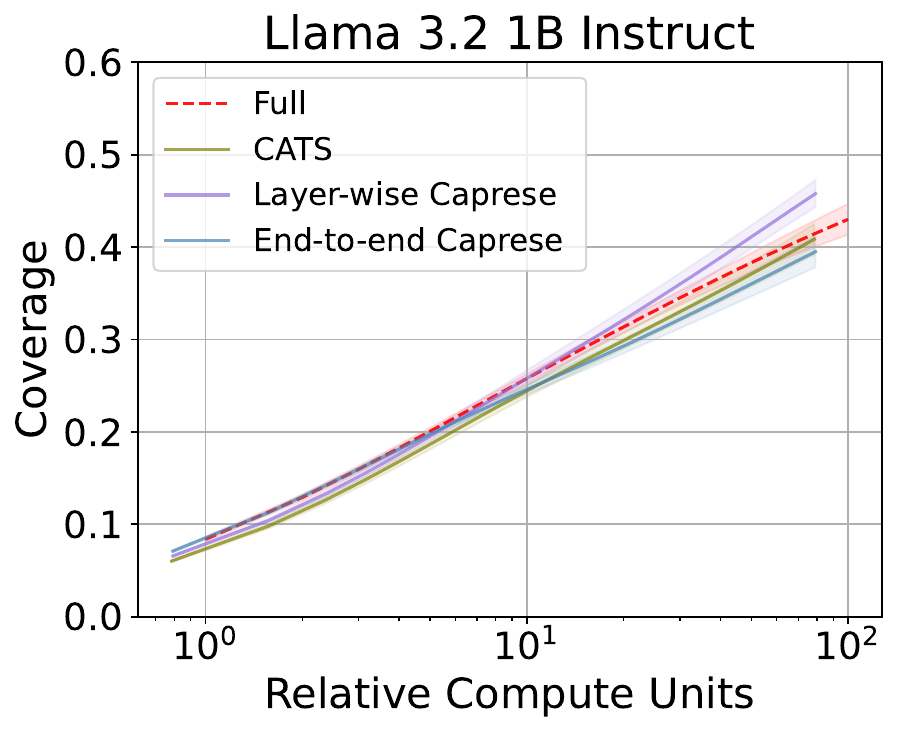}
\includegraphics[width=0.3\columnwidth]{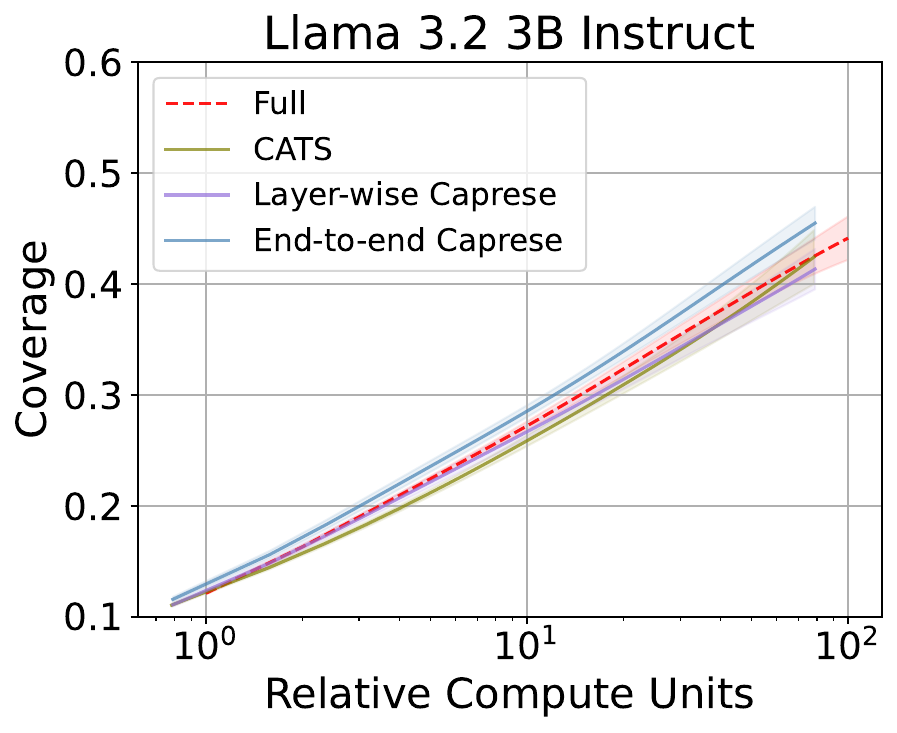}
\includegraphics[width=0.3\columnwidth]{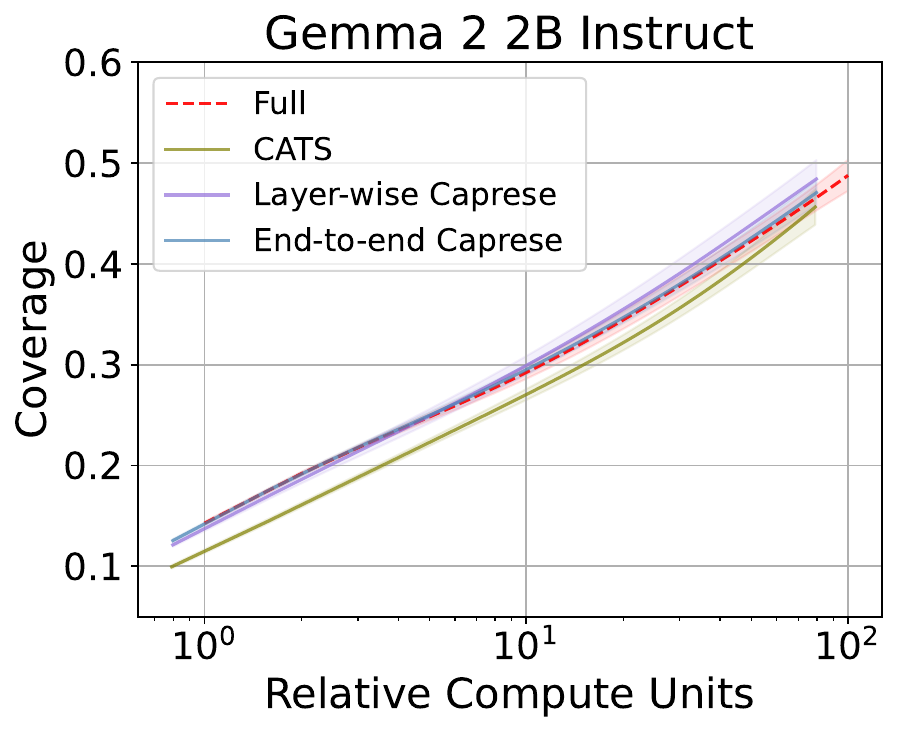}
\vspace{-0.1in}
\caption{Coverage and standard deviation of 140 samples from MATH as the number of generation attempts, $N$, scales. We define $\text{Relative Compute Units} = N \times A$ where $A$ is the fraction of total parameters activated per input. FF sparsity set to 50\%. Best viewed zoomed.}
\label{fig:coverage}
\end{center}
\vspace{-0.2in}
\end{figure}

Reselection is also possible with CATS but requires recomputing prefill for all tokens. The parameter to update is the hard thresholding parameter $\tau$, but since this requires finding the desired percentile of intermediate values in the FF block, we would need to access to all values, which likely cannot be fully stored in memory. In turn, prefill will need to be redone anytime we want to update $\tau$. Since $\tau$ represents a cutoff for a desired percentile (e.g., median), it is fairly robust to new observations. Given this and the lack of computational incentive, we focus primarily on reselection for \griffin{}.

\subsection{Scaling Best-of-$N$: Coverage}
\label{sec:coverage_experiments}

Now, we see the generalizability of \methodname{} on the second axis of test-time scaling: sampling multiple responses and selecting the best one, known as Best-of-$N$. 
Increasing the number of generations per prompt, $N$, increases the probability of generating a correct answer, even for non-thinking models (e.g., instruct models).
To evaluate, we find the coverage (Pass@$K$) on 140 samples from MATH. We use an oracle verifier to accurately assess the quality of the pool of generated responses. To combat the high variance when $K$ is close to $N$, we calculate the average coverage for $K=1, \dots, 100$ across 10 independent pools of 100 generations for each sample. \textit{Factoring in the saved compute, E2E \methodname{} is able to have similar or better coverage scaling compared to the original model}, as shown in Figure~\ref{fig:coverage}. Notably, E2E \methodname{} (\griffin{}) on Llama 3.2 3B Instruct improves Pass@$100$ by 7.0\% from the original model while using 15.8\% less compute.

\begin{table}[ht]
\begin{center}
\caption{Instruct models' 0-shot accuracies on mathematical reasoning (GSM8K and MATH) and language generation tasks (CoQA, QASPER, XSum, and CNN/DailyMail). FF sparsity is set to 50\%.}

\begin{tabular}{lcc|ccccc}
\toprule
Model & GSM8K & MATH & CoQA & QASPER & XSum & CNN/DailyMail \\
\midrule
\textit{Llama 3.2 1B Instruct}                    & 22.44	& 10.66	& 55.43		& 14.43 & 21.65 & 25.60 \\
\hdashline
\griffin{}              & 7.13	& 5.42 & 56.05	& 14.11 & 21.13 & 25.47 \\
Layer-wise \methodname{}    & 13.72	& 6.62 & 56.07	& 13.40 & 20.65 & 26.18\\
E2E \methodname{}    & \textbf{21.00}	& \textbf{8.44}	& 56.55	& 13.88 & 20.71  & 26.18 \\
\hdashline
CATS                            & 19.18	& 7.54 & 54.40	& 13.88 & 21.10 & 24.73 \\
Layer-wise \methodname{}                   & 18.65	& 8.28 & 55.58 	& 14.35 & 20.94 & 25.35 \\
E2E \methodname{}                   & \textbf{20.39}	& \textbf{9.04} & 56.12	& 13.75 & 20.40 & 25.56\\
\midrule

\textit{Llama 3.2 3B Instruct}                    & 51.55 & 14.32	& 63.95 & 12.45 & 23.22 & 26.20 \\
\hdashline
\griffin{}              & 28.96 & 10.98	& 64.52 & 12.52 & 22.09 & 25.49 \\
Layer-wise \methodname{}    & 40.18 & 13.70	& 64.33 & 11.60 & 21.56 & 25.90 \\
E2E \methodname{}    & \textbf{44.66}	& \textbf{16.96}	& 64.83	& 12.35 & 21.26 & 26.04 \\
\hdashline
CATS                            & 41.24 & 12.04 & 58.87 & 11.36 & 22.23 & 25.40 \\
Layer-wise \methodname{}                   & 45.49	& 13.62	& 60.53 & 12.26 & 22.50 & 25.90 \\
E2E \methodname{}                   & \textbf{46.85}	& \textbf{14.40}	& 61.72	& 12.36 & 21.58 & 25.40 \\
\midrule

\textit{Llama 3.1 8B Instruct}                    & 53.98	& 13.94	& 63.88	& 15.16 & 21.97 & 25.98 \\
\hdashline
\griffin{}              & 20.47	& 6.16	& 63.37   & 12.63 & 21.53 & 25.89 \\
Layer-wise \methodname{}    & 37.60	& 9.72	& 65.05	& 14.21 & 21.92 & 26.31 \\
E2E \methodname{}    & \textbf{51.40}	& \textbf{12.64}	& 65.50	& 15.35 & 22.05 & 26.42 \\
\hdashline
CATS                            & 50.95	& 11.80 & 58.85 & 12.26 & 21.43 & 25.67 \\
Layer-wise \methodname{}                  & 51.93	& 13.66	& 63.92    & 14.34 & 22.58 & 25.79 \\
E2E \methodname{}                   & \textbf{58.00}	& \textbf{13.88}	& 64.27	& 14.42 & 22.08 & 26.13 \\
\midrule

\textit{Gemma 2 2B Instruct}                      & 51.02 & 16.06	& 63.77 & 10.96 & 22.17 & 26.01\\
\hdashline
\griffin{}              & 33.74 & 11.32	& 63.28 & 11.07 & 18.27 & 22.24 \\
Layer-wise \methodname{}    & 42.53 & 12.32	& 63.77 & 10.75 & 21.63 & 26.37 \\
E2E \methodname{}    & \textbf{48.14}	& \textbf{13.70}	& 63.37	& 11.05 & 22.48 & 27.16 \\
\hdashline
CATS                            & 34.42	& 10.56	& 61.53	& 10.11 & 21.97 & 26.46 \\
Layer-wise \methodname{}                   & \textbf{46.32}	& 13.90	& 61.92	& 10.82 & 22.10 & 26.23 \\
E2E \methodname{}                   & 46.17	& \textbf{14.16}	& 63.92	& 11.03 & 22.15 & 26.52 \\

\midrule

\textit{Gemma 2 9B Instruct}                      & 78.17	& 27.64	& 63.78	& 9.91 & 23.98 & 26.46 \\
\hdashline
\griffin{}              & 59.21	& 25.22	& 63.82	& 10.14 & 24.66 & 26.77 \\
Layer-wise \methodname{}    & \textbf{76.72}	& \textbf{25.84}	& 64.20	& 9.82 & 24.88 & 26.88\\
E2E \methodname{}    & 76.65 & 25.30 & 64.42	& 9.92 & 24.89 & 25.47\\
\hdashline
CATS                           & 76.50	& 27.32 & 64.37	& 9.78 & 24.74 & 26.64 \\
Layer-wise \methodname{}                   & \textbf{77.18} & \textbf{28.16} & 64.52	& 10.46 & 24.54 & 26.45 \\
E2E \methodname{}                   & \textbf{77.18}	& 28.00	& 64.87	& 10.02 & 24.65 & 26.33 \\

\bottomrule
\end{tabular}
\label{tab:instruct_results_small}
\end{center}
\vspace{-0.1in}
\end{table}

\subsection{Instruct Models}
\label{sec:instruct_experiments}

Next, we look into the performance of \methodname{} in the absence of inference scaling methods on instruct LLMs whcih are primarily tailored for language tasks. We show \methodname{} is able to preserve math performance without sacrificing quality on language tasks like question answering. We test Llama 3 \cite{dubey2024llama} and Gemma 2 \cite{team2024gemma2} models on 0-shot GSM8K \cite{cobbe2021gsm8k}, MATH \cite{hendrycks2021measuring}, CoQA \cite{reddy2019coqa}, QASPER \cite{dasigi2021dataset}, XSum \cite{narayan2018don}, and CNN/DailyMail \cite{hermann2015teaching, see-etal-2017-get}. We use CoT prompts for math tasks. 

Table~\ref{tab:instruct_results_small} shows \textit{\methodname{} preserves most if not all of the math capabilities in the original models without damaging performance on the language tasks, despite the distillation dataset being all math.} In most cases, CATS and \griffin{} severely harm GSM8K and MATH accuracy, but \methodname{} effectively recovers the lost performance. E2E \methodname{} is the most performative in the majority of math scenarios. 
\methodname{}'s performance is consistent at different sparsity levels and $r$ (Appendix~\ref{app:rank_v_sparsity}). 
All methods have little impact on the accuracy of these tasks (the main purpose of these tasks is to show no degradation in language-related cases). 

\subsection{Efficiency}
\label{sec:efficiency}

\methodname{} reduces generation latency (Table~\ref{tab:griffin_efficiency}). \methodname{} cuts total latency and time to next token by >16\% for the Qwen 2.5 14B architecture. Moreover, the latency differences between \griffin{} and \methodname{} are exceptionally small, suggesting that \methodname{} has \textit{minimal overhead}. We also include latencies of \griffin{} paired with LoRA weights, which although faster than the base model, negates >40\% of the time savings that \griffin{} has provided. This highlights the efficiency suboptimality of LoRA during inference. Metrics were collected on an NVIDIA L40 GPU using BF16 precision.

\begin{table}[t]
\caption{End-to-end generation latency (s) and average time to next token (ms) for Qwen 2.5 14B. For the ``Setup'' column, $P+G$ indicates input and generation lengths of $P$ and $G$ tokens, respectively. As before, \griffin{} is applied to the first half of the model, sparsity is 50\%, and $r=256$.}
\begin{center}
\begin{tabular}{c|cccc|cccc}
\toprule
\multirow{2}{*}{Setup} & 
\multicolumn{4}{c|}{End-to-end Latency (s)} &
\multicolumn{4}{c}{Avg. Time to Next Token (ms)} \\
&  Full & \griffin{} & LoRA & \methodname{} &  Full & \griffin{} & LoRA & \methodname{}  \\
\midrule
2048+256 & 10.5 & 8.7 & 9.5 & 8.7 & 41 & 34 & 37 & 34\\ 
2048+2048 & 84.4 & 70.3 & 76.5 & 70.4 & 41 & 34 & 37 & 34\\ 
2048+8192 & 344.9 & 287.7 & 312.5 & 288.8 & 42 & 35 & 38 & 35 \\ 

\bottomrule
\end{tabular}
\end{center}
\label{tab:griffin_efficiency}
\end{table}

\subsubsection{Bonus: Effect on Natural Response Length}
\label{sec:resp_length}

\begin{figure}[t]
\begin{center}
\includegraphics[width=0.32\columnwidth]{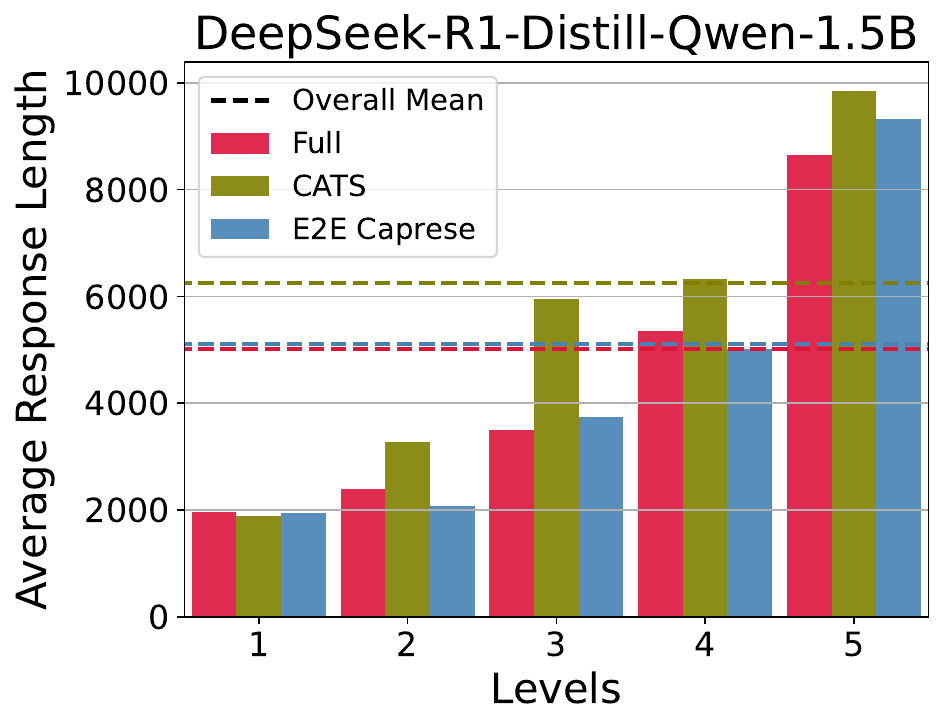}
\includegraphics[width=0.32\columnwidth]{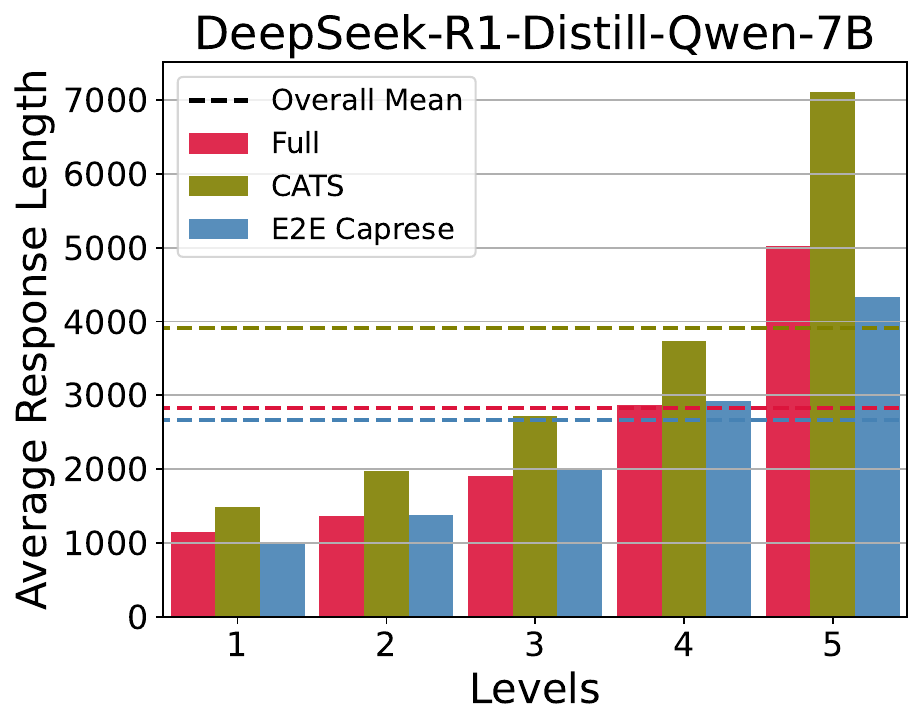}
\includegraphics[width=0.32\columnwidth]{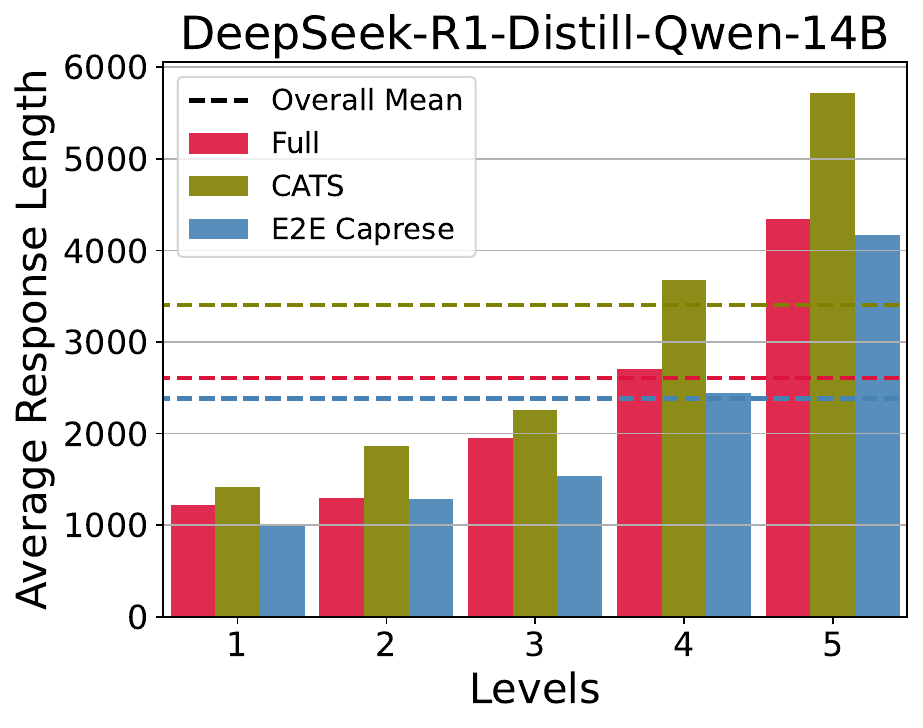}
\vspace{-0.12in}
\caption{Average number of response tokens for MATH-500 queries with increasing problem difficulty. The global averages are indicated by the dashed lines. Sparsity is set at 50\%.}
\label{fig:resp_lengths}
\end{center}
\vspace{-0.1in}
\end{figure}

\methodname{} implicitly encourages brevity, often even producing shorter responses than from the full thinking model. Intriguingly, this behavior arises despite any enforcement or regularization on response lengths anywhere during training or inference. Shown in Figure~\ref{fig:resp_lengths}, the shortest mean response length for all MATH-500 problem difficulties is either the full model or \methodname{}. CATS consistently outputs the longest responses, averaging $\sim$1K more across every sample. With increasing difficulty, all models and methods naturally allocate more inference tokens towards the output. Given CATS and \methodname{} (CATS) achieve similar MATH-500 accuracies to the full DS-Qwen 7B and 14B models, the ability of \methodname{} to cut down the lengthy responses of CATS down to the response lengths of the original model or shorter without compromising performance is a direct memory and latency benefit. This brevity of \methodname{} is more pronounced in larger models, proportionally, from a 5.8\% reduction in tokens compared to the full DS-Qwen 7B model to an 8.5\% reduction compared to the full DS-Qwen 14B model. See Appendix~\ref{app:griffin_resp_lengths} for observations with \griffin{}.

\section{Conclusion}
\label{sec:conclusion}

To combat the inefficiency of the long and brittle generation process associated with reasoning, we introduce \methodname{}, a performant and efficient method with strong reasoning and language capabilities that is compatible with a broad class of efficient FF algorithms. In the future, it would be interesting to evaluate its benefit in more reasoning domains, explore input-dependent low-rank layers, and investigate the knowledge contained in these low-rank layers. \methodname{}, along with future developments, pushes towards the long-term goal of degradation-free efficient LLM inference.

\section*{Acknowledgements}
The work of H. Dong is supported in part by the Wei Shen and Xuehong Zhang Presidential Fellowship at Carnegie Mellon University.

\bibliographystyle{alphaabbr}
\bibliography{bibliography.bib}

@article{vaswani2017attention,
  title={Attention is all you need},
  author={Vaswani, Ashish and Shazeer, Noam and Parmar, Niki and Uszkoreit, Jakob and Jones, Llion and Gomez, Aidan N and Kaiser, {\L}ukasz and Polosukhin, Illia},
  journal={Advances in neural information processing systems},
  volume={30},
  year={2017}
}

@article{sun2023simple,
  title={A Simple and Effective Pruning Approach for Large Language Models},
  author={Sun, Mingjie and Liu, Zhuang and Bair, Anna and Kolter, J Zico},
  journal={arXiv preprint arXiv:2306.11695},
  year={2023}
}

@misc{zhang2022opt,
      title={OPT: Open Pre-trained Transformer Language Models}, 
      author={Susan Zhang and Stephen Roller and Naman Goyal and Mikel Artetxe and Moya Chen and Shuohui Chen and Christopher Dewan and Mona Diab and Xian Li and Xi Victoria Lin and Todor Mihaylov and Myle Ott and Sam Shleifer and Kurt Shuster and Daniel Simig and Punit Singh Koura and Anjali Sridhar and Tianlu Wang and Luke Zettlemoyer},
      year={2022},
      eprint={2205.01068},
      archivePrefix={arXiv},
      primaryClass={cs.CL}
}

@article{frantar2023massive,
  title={Massive language models can be accurately pruned in one-shot},
  author={Frantar, Elias and Alistarh, Dan},
  journal={arXiv preprint arXiv:2301.00774},
  year={2023}
}

@article{zhang2021moefication,
  title={Moefication: Transformer feed-forward layers are mixtures of experts},
  author={Zhang, Zhengyan and Lin, Yankai and Liu, Zhiyuan and Li, Peng and Sun, Maosong and Zhou, Jie},
  journal={arXiv preprint arXiv:2110.01786},
  year={2021}
}

@inproceedings{li2022lazy,
  title={The Lazy Neuron Phenomenon: On Emergence of Activation Sparsity in Transformers},
  author={Li, Zonglin and You, Chong and Bhojanapalli, Srinadh and Li, Daliang and Rawat, Ankit Singh and Reddi, Sashank J and Ye, Ke and Chern, Felix and Yu, Felix and Guo, Ruiqi and others},
  booktitle={The Eleventh International Conference on Learning Representations},
  year={2022}
}

@inproceedings{liu2023deja,
  title={Deja vu: Contextual sparsity for efficient llms at inference time},
  author={Liu, Zichang and Wang, Jue and Dao, Tri and Zhou, Tianyi and Yuan, Binhang and Song, Zhao and Shrivastava, Anshumali and Zhang, Ce and Tian, Yuandong and Re, Christopher and others},
  booktitle={International Conference on Machine Learning},
  pages={22137--22176},
  year={2023},
  organization={PMLR}
}

@inproceedings{see-etal-2017-get,
    title = "Get To The Point: Summarization with Pointer-Generator Networks",
    author = "See, Abigail  and
      Liu, Peter J.  and
      Manning, Christopher D.",
    booktitle = "Proceedings of the 55th Annual Meeting of the Association for Computational Linguistics (Volume 1: Long Papers)",
    month = jul,
    year = "2017",
    address = "Vancouver, Canada",
    publisher = "Association for Computational Linguistics",
    url = "https://www.aclweb.org/anthology/P17-1099",
    doi = "10.18653/v1/P17-1099",
    pages = "1073--1083",
}

@inproceedings{hermann2015teaching,
  author={Karl Moritz Hermann and Tomás Kociský and Edward Grefenstette and Lasse Espeholt and Will Kay and Mustafa Suleyman and Phil Blunsom},
  title={Teaching Machines to Read and Comprehend},
  year={2015},
  cdate={1420070400000},
  pages={1693-1701},
  booktitle={NIPS}
}

@article{narayan2018don,
  title={Don't give me the details, just the summary! topic-aware convolutional neural networks for extreme summarization},
  author={Narayan, Shashi and Cohen, Shay B and Lapata, Mirella},
  journal={arXiv preprint arXiv:1808.08745},
  year={2018}
}

@article{dettmers2022llm,
  title={Llm. int8 (): 8-bit matrix multiplication for transformers at scale},
  author={Dettmers, Tim and Lewis, Mike and Belkada, Younes and Zettlemoyer, Luke},
  journal={arXiv preprint arXiv:2208.07339},
  year={2022}
}

@article{jacobs1991adaptive,
  title={Adaptive mixtures of local experts},
  author={Jacobs, Robert A and Jordan, Michael I and Nowlan, Steven J and Hinton, Geoffrey E},
  journal={Neural computation},
  volume={3},
  number={1},
  pages={79--87},
  year={1991},
  publisher={MIT Press}
}

@inproceedings{dong2023towards,
  title={Towards Structured Sparsity in Transformers for Efficient Inference},
  author={Dong, Harry and Chen, Beidi and Chi, Yuejie},
  booktitle={Workshop on Efficient Systems for Foundation Models@ ICML2023},
  year={2023}
}

@article{lecun1989optimal,
  title={Optimal brain damage},
  author={LeCun, Yann and Denker, John and Solla, Sara},
  journal={Advances in neural information processing systems},
  volume={2},
  year={1989}
}

@article{ma2023llm,
  title={Llm-pruner: On the structural pruning of large language models},
  author={Ma, Xinyin and Fang, Gongfan and Wang, Xinchao},
  journal={Advances in neural information processing systems},
  volume={36},
  pages={21702--21720},
  year={2023}
}

@article{geva2020transformer,
  title={Transformer feed-forward layers are key-value memories},
  author={Geva, Mor and Schuster, Roei and Berant, Jonathan and Levy, Omer},
  journal={arXiv preprint arXiv:2012.14913},
  year={2020}
}

@article{dasigi2021dataset,
  title={A dataset of information-seeking questions and answers anchored in research papers},
  author={Dasigi, Pradeep and Lo, Kyle and Beltagy, Iz and Cohan, Arman and Smith, Noah A and Gardner, Matt},
  journal={arXiv preprint arXiv:2105.03011},
  year={2021}
}

@article{reddy2019coqa,
  title={Coqa: A conversational question answering challenge},
  author={Reddy, Siva and Chen, Danqi and Manning, Christopher D},
  journal={Transactions of the Association for Computational Linguistics},
  volume={7},
  pages={249--266},
  year={2019},
  publisher={MIT Press One Rogers Street, Cambridge, MA 02142-1209, USA journals-info~…}
}

@article{kaplan2020scaling,
  title={Scaling laws for neural language models},
  author={Kaplan, Jared and McCandlish, Sam and Henighan, Tom and Brown, Tom B and Chess, Benjamin and Child, Rewon and Gray, Scott and Radford, Alec and Wu, Jeffrey and Amodei, Dario},
  journal={arXiv preprint arXiv:2001.08361},
  year={2020}
}

@article{hoffmann2022training,
  title={Training compute-optimal large language models},
  author={Hoffmann, Jordan and Borgeaud, Sebastian and Mensch, Arthur and Buchatskaya, Elena and Cai, Trevor and Rutherford, Eliza and Casas, Diego de Las and Hendricks, Lisa Anne and Welbl, Johannes and Clark, Aidan and others},
  journal={arXiv preprint arXiv:2203.15556},
  year={2022}
}

@article{team2024gemma2,
  title={Gemma 2: Improving open language models at a practical size},
  author={Team, Gemma and Riviere, Morgane and Pathak, Shreya and Sessa, Pier Giuseppe and Hardin, Cassidy and Bhupatiraju, Surya and Hussenot, L{\'e}onard and Mesnard, Thomas and Shahriari, Bobak and Ram{\'e}, Alexandre and others},
  journal={arXiv preprint arXiv:2408.00118},
  year={2024}
}

@article{dubey2024llama,
  title={The llama 3 herd of models},
  author={Dubey, Abhimanyu and Jauhri, Abhinav and Pandey, Abhinav and Kadian, Abhishek and Al-Dahle, Ahmad and Letman, Aiesha and Mathur, Akhil and Schelten, Alan and Yang, Amy and Fan, Angela and others},
  journal={arXiv preprint arXiv:2407.21783},
  year={2024}
}

@inproceedings{dong2024get,
  title={Get More with LESS: Synthesizing Recurrence with KV Cache Compression for Efficient LLM Inference},
  author={Dong, Harry and Yang, Xinyu and Zhang, Zhenyu and Wang, Zhangyang and Chi, Yuejie and Chen, Beidi},
  booktitle={Forty-first International Conference on Machine Learning},
  year={2024},
}

@inproceedings{dong2024prompt,
  title={Prompt-prompted Adaptive Structured Pruning for Efficient LLM Generation},
  author={Dong, Harry and Chen, Beidi and Chi, Yuejie},
  booktitle={First Conference on Language Modeling},
  year={2024}
}

@article{brown2024large,
  title={Large language monkeys: Scaling inference compute with repeated sampling},
  author={Brown, Bradley and Juravsky, Jordan and Ehrlich, Ryan and Clark, Ronald and Le, Quoc V and R{\'e}, Christopher and Mirhoseini, Azalia},
  journal={arXiv preprint arXiv:2407.21787},
  year={2024}
}

@article{snell2024scaling,
  title={Scaling llm test-time compute optimally can be more effective than scaling model parameters},
  author={Snell, Charlie and Lee, Jaehoon and Xu, Kelvin and Kumar, Aviral},
  journal={arXiv preprint arXiv:2408.03314},
  year={2024}
}

@article{wu2024empirical,
  title={An empirical analysis of compute-optimal inference for problem-solving with language models},
  author={Wu, Yangzhen and Sun, Zhiqing and Li, Shanda and Welleck, Sean and Yang, Yiming},
  journal={arXiv preprint arXiv:2408.00724},
  year={2024}
}

@article{guo2025deepseek,
  title={Deepseek-r1: Incentivizing reasoning capability in llms via reinforcement learning},
  author={Guo, Daya and Yang, Dejian and Zhang, Haowei and Song, Junxiao and Zhang, Ruoyu and Xu, Runxin and Zhu, Qihao and Ma, Shirong and Wang, Peiyi and Bi, Xiao and others},
  journal={arXiv preprint arXiv:2501.12948},
  year={2025}
}

@article{zhou2024sirius,
  title={Sirius: Contextual sparsity with correction for efficient llms},
  author={Zhou, Yang and Chen, Zhuoming and Xu, Zhaozhuo and Lin, Victoria and Chen, Beidi},
  journal={arXiv preprint arXiv:2409.03856},
  year={2024}
}

@article{dong2024towards,
  title={Towards Low-bit Communication for Tensor Parallel LLM Inference},
  author={Dong, Harry and Johnson, Tyler and Cho, Minsik and Soroush, Emad},
  journal={arXiv preprint arXiv:2411.07942},
  year={2024}
}

@article{manvi2024adaptive,
  title={Adaptive inference-time compute: Llms can predict if they can do better, even mid-generation},
  author={Manvi, Rohin and Singh, Anikait and Ermon, Stefano},
  journal={arXiv preprint arXiv:2410.02725},
  year={2024}
}

@article{sun2024fast,
  title={Fast best-of-n decoding via speculative rejection},
  author={Sun, Hanshi and Haider, Momin and Zhang, Ruiqi and Yang, Huitao and Qiu, Jiahao and Yin, Ming and Wang, Mengdi and Bartlett, Peter and Zanette, Andrea},
  journal={arXiv preprint arXiv:2410.20290},
  year={2024}
}

@article{wei2022chain,
  title={Chain-of-thought prompting elicits reasoning in large language models},
  author={Wei, Jason and Wang, Xuezhi and Schuurmans, Dale and Bosma, Maarten and Xia, Fei and Chi, Ed and Le, Quoc V and Zhou, Denny and others},
  journal={Advances in neural information processing systems},
  volume={35},
  pages={24824--24837},
  year={2022}
}

@article{cobbe2021gsm8k,
  title={Training Verifiers to Solve Math Word Problems},
  author={Cobbe, Karl and Kosaraju, Vineet and Bavarian, Mohammad and Chen, Mark and Jun, Heewoo and Kaiser, Lukasz and Plappert, Matthias and Tworek, Jerry and Hilton, Jacob and Nakano, Reiichiro and Hesse, Christopher and Schulman, John},
  journal={arXiv preprint arXiv:2110.14168},
  year={2021}
}

@article{hendrycks2021measuring,
  title={Measuring mathematical problem solving with the math dataset},
  author={Hendrycks, Dan and Burns, Collin and Kadavath, Saurav and Arora, Akul and Basart, Steven and Tang, Eric and Song, Dawn and Steinhardt, Jacob},
  journal={arXiv preprint arXiv:2103.03874},
  year={2021}
}

@article{lee2024cats,
  title={Cats: Contextually-aware thresholding for sparsity in large language models},
  author={Lee, Donghyun and Lee, Je-Yong and Zhang, Genghan and Tiwari, Mo and Mirhoseini, Azalia},
  journal={arXiv preprint arXiv:2404.08763},
  year={2024}
}

@article{arora2025training,
  title={Training Language Models to Reason Efficiently},
  author={Arora, Daman and Zanette, Andrea},
  journal={arXiv preprint arXiv:2502.04463},
  year={2025}
}

@misc{aytes2025sketch,
      title={Sketch-of-Thought: Efficient LLM Reasoning with Adaptive Cognitive-Inspired Sketching}, 
      author={Simon A. Aytes and Jinheon Baek and Sung Ju Hwang},
      year={2025},
      eprint={2503.05179},
      archivePrefix={arXiv},
      primaryClass={cs.CL},
      url={https://arxiv.org/abs/2503.05179}, 
}

@inproceedings{renze2024benefits,
  title={The benefits of a concise chain of thought on problem-solving in large language models},
  author={Renze, Matthew and Guven, Erhan},
  booktitle={2024 2nd International Conference on Foundation and Large Language Models (FLLM)},
  pages={476--483},
  year={2024},
  organization={IEEE}
}

@article{nayab2024concise,
  title={Concise thoughts: Impact of output length on llm reasoning and cost},
  author={Nayab, Sania and Rossolini, Giulio and Simoni, Marco and Saracino, Andrea and Buttazzo, Giorgio and Manes, Nicolamaria and Giacomelli, Fabrizio},
  journal={arXiv preprint arXiv:2407.19825},
  year={2024}
}

@article{dai2024deepseekmoe,
  title={Deepseekmoe: Towards ultimate expert specialization in mixture-of-experts language models},
  author={Dai, Damai and Deng, Chengqi and Zhao, Chenggang and Xu, RX and Gao, Huazuo and Chen, Deli and Li, Jiashi and Zeng, Wangding and Yu, Xingkai and Wu, Yu and others},
  journal={arXiv preprint arXiv:2401.06066},
  year={2024}
}

@article{xu2025chain,
  title={Chain of Draft: Thinking Faster by Writing Less},
  author={Xu, Silei and Xie, Wenhao and Zhao, Lingxiao and He, Pengcheng},
  journal={arXiv preprint arXiv:2502.18600},
  year={2025}
}

@inproceedings{qu2025optimizing,
  title={Optimizing Test-Time Compute via Meta Reinforcement Finetuning},
  author={Qu, Yuxiao and Yang, Matthew YR and Setlur, Amrith and Tunstall, Lewis and Beeching, Edward Emanuel and Salakhutdinov, Ruslan and Kumar, Aviral},
  booktitle={Workshop on Reasoning and Planning for Large Language Models},
  year={2025}
}

@article{yang2024qwen2,
  title={Qwen2. 5 technical report},
  author={Yang, An and Yang, Baosong and Zhang, Beichen and Hui, Binyuan and Zheng, Bo and Yu, Bowen and Li, Chengyuan and Liu, Dayiheng and Huang, Fei and Wei, Haoran and others},
  journal={arXiv preprint arXiv:2412.15115},
  year={2024}
}

@misc{openr1,
    title = {Open R1: A fully open reproduction of DeepSeek-R1},
    url = {https://github.com/huggingface/open-r1},
    author = {Hugging Face},
    month = {January},
    year = {2025}
}

@article{lightman2023lets,
      title={Let's Verify Step by Step}, 
      author={Lightman, Hunter and Kosaraju, Vineet and Burda, Yura and Edwards, Harri and Baker, Bowen and Lee, Teddy and Leike, Jan and Schulman, John and Sutskever, Ilya and Cobbe, Karl},
      journal={arXiv preprint arXiv:2305.20050},
      year={2023}
}

@article{chen2023accelerating,
  title={Accelerating large language model decoding with speculative sampling},
  author={Chen, Charlie and Borgeaud, Sebastian and Irving, Geoffrey and Lespiau, Jean-Baptiste and Sifre, Laurent and Jumper, John},
  journal={arXiv preprint arXiv:2302.01318},
  year={2023}
}

@inproceedings{leviathan2023fast,
  title={Fast inference from transformers via speculative decoding},
  author={Leviathan, Yaniv and Kalman, Matan and Matias, Yossi},
  booktitle={International Conference on Machine Learning},
  pages={19274--19286},
  year={2023},
  organization={PMLR}
}

@article{hu2022lora,
  title={Lora: Low-rank adaptation of large language models.},
  author={Hu, Edward J and Shen, Yelong and Wallis, Phillip and Allen-Zhu, Zeyuan and Li, Yuanzhi and Wang, Shean and Wang, Lu and Chen, Weizhu and others},
  journal={ICLR},
  volume={1},
  number={2},
  pages={3},
  year={2022}
}

@misc{aime,
      title={{AIME} Problems and Solutions},
      author={{AIME}},
      year={2025},
      url={https://artofproblemsolving.com/wiki/index.php/AIME_Problems_and_Solutions}
}

@misc{amc23,
      title={American Mathematics Competition}, 
      author={AMC},
      year={2023},
      accessed={2025},
      url={https://maa.org/student-programs/amc/}, 
}

@misc{brumo25,
      title={Brown university math olympiad 2025}, 
      author={BRUMO},
      year={2025},
      accessed={2025},
      url={https://www.brumo.org/}, 
}

@article{yang2025qwen3,
  title={Qwen3 technical report},
  author={Yang, An and Li, Anfeng and Yang, Baosong and Zhang, Beichen and Hui, Binyuan and Zheng, Bo and Yu, Bowen and Gao, Chang and Huang, Chengen and Lv, Chenxu and others},
  journal={arXiv preprint arXiv:2505.09388},
  year={2025}
}

@article{muennighoff2025s1,
  title={s1: Simple test-time scaling},
  author={Muennighoff, Niklas and Yang, Zitong and Shi, Weijia and Li, Xiang Lisa and Fei-Fei, Li and Hajishirzi, Hannaneh and Zettlemoyer, Luke and Liang, Percy and Cand{\`e}s, Emmanuel and Hashimoto, Tatsunori},
  journal={arXiv preprint arXiv:2501.19393},
  year={2025}
}

@inproceedings{rein2024gpqa,
  title={Gpqa: A graduate-level google-proof q\&a benchmark},
  author={Rein, David and Hou, Betty Li and Stickland, Asa Cooper and Petty, Jackson and Pang, Richard Yuanzhe and Dirani, Julien and Michael, Julian and Bowman, Samuel R},
  booktitle={First Conference on Language Modeling},
  year={2024}
}

@article{dong2025generalized,
  title={Generalized Parallel Scaling with Interdependent Generations},
  author={Dong, Harry and Brandfonbrener, David and Helenowski, Eryk and He, Yun and Kumar, Mrinal and Fang, Han and Chi, Yuejie and Sankararaman, Karthik Abinav},
  journal={arXiv preprint arXiv:2510.01143},
  year={2025}
}

@article{dettmers2023qlora,
  title={Qlora: Efficient finetuning of quantized llms},
  author={Dettmers, Tim and Pagnoni, Artidoro and Holtzman, Ari and Zettlemoyer, Luke},
  journal={Advances in neural information processing systems},
  volume={36},
  pages={10088--10115},
  year={2023}
}

@article{dettmers2022gpt3,
  title={Gpt3. int8 (): 8-bit matrix multiplication for transformers at scale},
  author={Dettmers, Tim and Lewis, Mike and Belkada, Younes and Zettlemoyer, Luke},
  journal={Advances in neural information processing systems},
  volume={35},
  pages={30318--30332},
  year={2022}
}

@article{lin2024awq,
  title={Awq: Activation-aware weight quantization for on-device llm compression and acceleration},
  author={Lin, Ji and Tang, Jiaming and Tang, Haotian and Yang, Shang and Chen, Wei-Ming and Wang, Wei-Chen and Xiao, Guangxuan and Dang, Xingyu and Gan, Chuang and Han, Song},
  journal={Proceedings of machine learning and systems},
  volume={6},
  pages={87--100},
  year={2024}
}

@article{sadhukhan2026stem,
  title={STEM: Scaling Transformers with Embedding Modules},
  author={Sadhukhan, Ranajoy and Cao, Sheng and Dong, Harry and Zhao, Changsheng and Purpura-Pontoniere, Attiano and Tian, Yuandong and Liu, Zechun and Chen, Beidi},
  journal={arXiv preprint arXiv:2601.10639},
  year={2026}
}

\newpage

\appendix

\newpage
\section{Effect of Rank \& Sparsity}
\label{app:rank_v_sparsity}

\begin{figure}[h]
\begin{center}
\includegraphics[width=0.45\columnwidth]{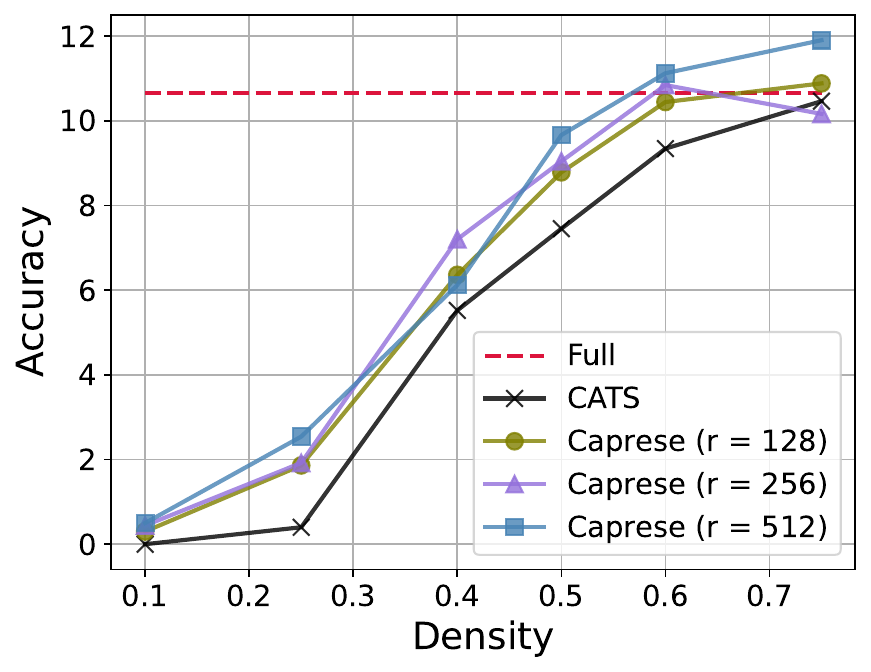}
\caption{Llama 3.2 1B Instruct's performance on MATH with varying densities of CATS and ranks in \methodname{} with end-to-end training.}
\label{fig:rank_v_sparsity}
\end{center}
\end{figure}

Here, we ablate the relationship between varying ranks in \methodname{} and sparsity levels in CATS. Using Llama 3.2 1B Instruct, we show the test performance of MATH in Figure~\ref{fig:rank_v_sparsity}. The same training procedure and data are used as outlined in Section~\ref{sec:method}. For all ablated ranks, \methodname{} consistently outperforms pure CATS by a large margin when CATS performs poorly relative to the full model. With a couple of exceptions (perhaps due to randomness in the generation process), greater performance is correlated with higher rank.

\section{\methodname{} Parameters}
\label{app:parameters}

\begin{figure}[h]
\begin{center}
\includegraphics[width=0.42\columnwidth]{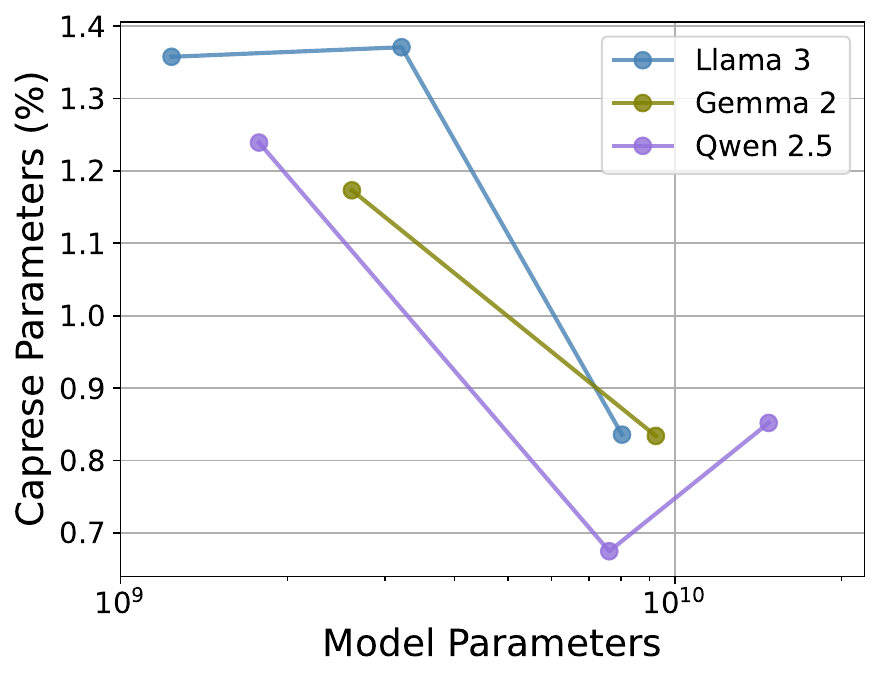}
\includegraphics[width=0.42\columnwidth]{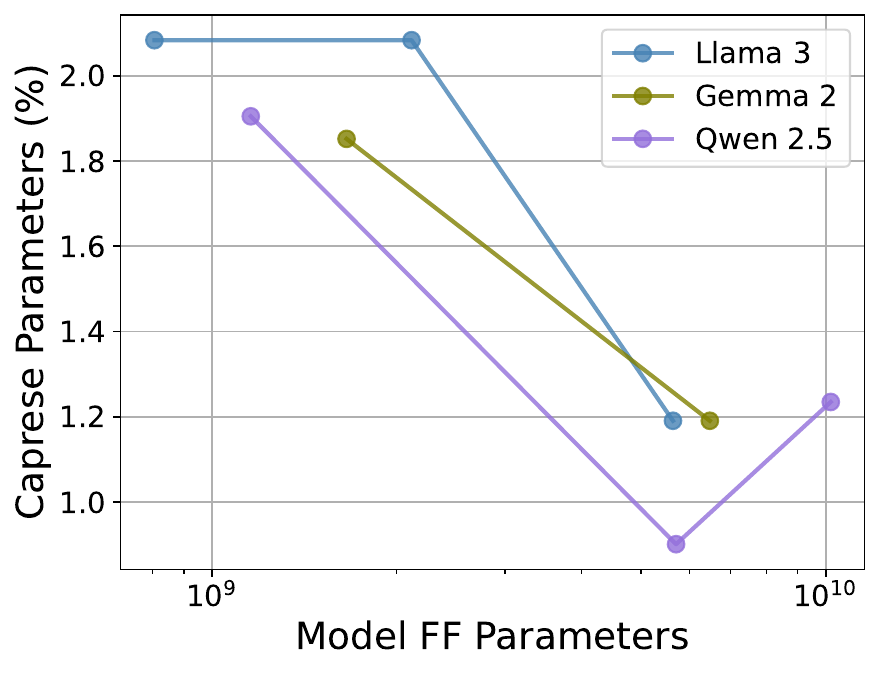}
\caption{The percent of new parameters that \methodname{} ($r=256$) adds relative to the entire model (left) and relative to only the FF parameters (right) for the Llama 3, Gemma 2, and Qwen 2.5 model families.}
\label{fig:params}
\end{center}
\end{figure}

\methodname{} has a tiny parameter footprint. In Figure~\ref{fig:params}, we see that \methodname{} adds roughly 1\% new parameters relative to the full model with a trend downwards as model size increases.

\newpage
\section{LoRA Performance}
\label{app:lora}

We report the performance of using LoRA in place of \methodname{} while keeping parameter count constant in Table~\ref{tab:instruct_results_lora}. \methodname{} achieves the highest accuracy in slightly more cases than LoRA, but as per our discussion on LoRA in Section~\ref{sec:lora} and efficiency analysis in Section~\ref{sec:efficiency}, LoRA's main benefit is efficient training, not efficient inference.

\begin{table}[h]
\caption{0-shot accuracies on mathematical reasoning (GSM8K and MATH) and language generation tasks (CoQA, QASPER, XSum, and CNN/DailyMail) with LoRA. FF sparsity is set to 50\%, $r = 256$, and LoRA ranks are set to match the number of parameters that \methodname{} adds.}
\vspace{-0.1in}
\begin{center}
\begin{tabular}{lcc|cccc}
\toprule
Model & GSM8K & MATH & CoQA & QASPER & XSum & CNN/DailyMail \\
\midrule
\textit{Llama 3.2 1B Instruct}                    & 22.44	& 10.66	& 55.43		& 14.43 & 21.65 & 25.60 \\
\hdashline
\griffin{}              & 7.13	& 5.42 & 56.05	& 14.11 & 21.13 & 25.47 \\
Layer-wise LoRA & 8.11 & 5.60 & 56.00 & 14.65 & 21.08 & 25.51\\
E2E LoRA & 20.47 & \textbf{10.10} & 56.10 & 14.41  & 21.09 & 25.62 \\
Layer-wise \methodname{}    & 13.72	& 6.62 & 56.07	& 13.40 & 20.65 & 26.18\\
E2E \methodname{}    & \textbf{21.00}	& 8.44	& 56.55	& 13.88 & 20.71  & 26.18 \\
\midrule

\textit{Llama 3.2 3B Instruct}                    & 51.55 & 14.32	& 63.95 & 12.45 & 23.22 & 26.20 \\
\hdashline
\griffin{}              & 28.96 & 10.98	& 64.52 & 12.52 & 22.09 & 25.49 \\
Layer-wise LoRA & 26.46 & 11.28 & 63.88 & 12.55 & 22.12 & 25.96 \\
E2E LoRA & 42.91 & 16.50 & 64.67 & 12.63 & 21.81 & 26.05 \\
Layer-wise \methodname{}    & 40.18 & 13.70	& 64.33 & 11.60 & 21.56 & 25.90 \\
E2E \methodname{}    & \textbf{44.66}	& \textbf{16.96}	& 64.83	& 12.35 & 21.26 & 26.04 \\
\midrule

\textit{Gemma 2 2B Instruct}                      & 51.02 & 16.06	& 63.77 & 10.96 & 22.17 & 26.01\\
\hdashline
\griffin{}              & 33.74 & 11.32	& 63.28 & 11.07 & 18.27 & 22.24 \\
Layer-wise LoRA & 39.58 & 14.66 & 62.98 & 10.54 & 21.58 & 26.64 \\
E2E LoRA & 44.66 & \textbf{16.71} & 63.03 & 10.81 & 22.45 & 26.35\\
Layer-wise \methodname{}    & 42.53 & 12.32	& 63.77 & 10.75 & 21.63 & 26.37 \\
E2E \methodname{}    & \textbf{48.14}	& \textbf{13.70}	& 63.37	& 11.05 & 22.48 & 27.16 \\

\bottomrule
\end{tabular}
\end{center}
\label{tab:instruct_results_lora}
\vspace{-0.2in}
\end{table}

\section{\methodname{} with Quantization}
\label{app:quantization}

To demonstrate recovery generalizability beyond sparse FF methods as the backend, we also equip QLoRA \cite{dettmers2023qlora} with \methodname{}. Table~\ref{tab:quantization} shows that E2E \methodname{} is also able to nearly recover the performance degraded by quantization in both math and language tasks.

\begin{table}[h]
\caption{0-shot accuracies on mathematical reasoning (GSM8K and MATH) and language generation tasks (CoQA, QASPER) with QLoRA as the backend method. FF weights are quantized to 4 bits and $r = 256$.}
\vspace{-0.1in}
\begin{center}
\begin{tabular}{lcc|cc}
\toprule
Model & GSM8K & MATH & CoQA & QASPER \\
\midrule
\textit{Llama 3.2 1B Instruct}  & 22.44 & 10.66 & 55.43 & 14.43 \\
\hdashline
QLoRA   & 12.05 & 6.56 & 48.70 & 12.37 \\
Layer-wise \methodname{}    & 15.01 & 6.88 & 49.60 & 12.65 \\
E2E \methodname{}    & \textbf{21.61} & \textbf{9.98} & 53.52 & 13.54 \\
\midrule

\textit{Llama 3.2 3B Instruct}  & 51.55 & 14.32 & 63.95 & 12.45 \\
\hdashline
QLoRA   & 45.41 & 11.98 & 62.22 & 12.87 \\
Layer-wise \methodname{}    & 45.56 & 11.78 & 64.25 & 12.41 \\
E2E \methodname{}    & \textbf{48.82} & \textbf{14.14} & 64.05 & 12.26 \\
\bottomrule
\end{tabular}
\end{center}
\label{tab:quantization}
\vspace{-0.15in}
\end{table}




\section{\griffin{} Response Lengths}
\label{app:griffin_resp_lengths}

\begin{figure}[h]
\begin{center}
\includegraphics[width=0.31\columnwidth]{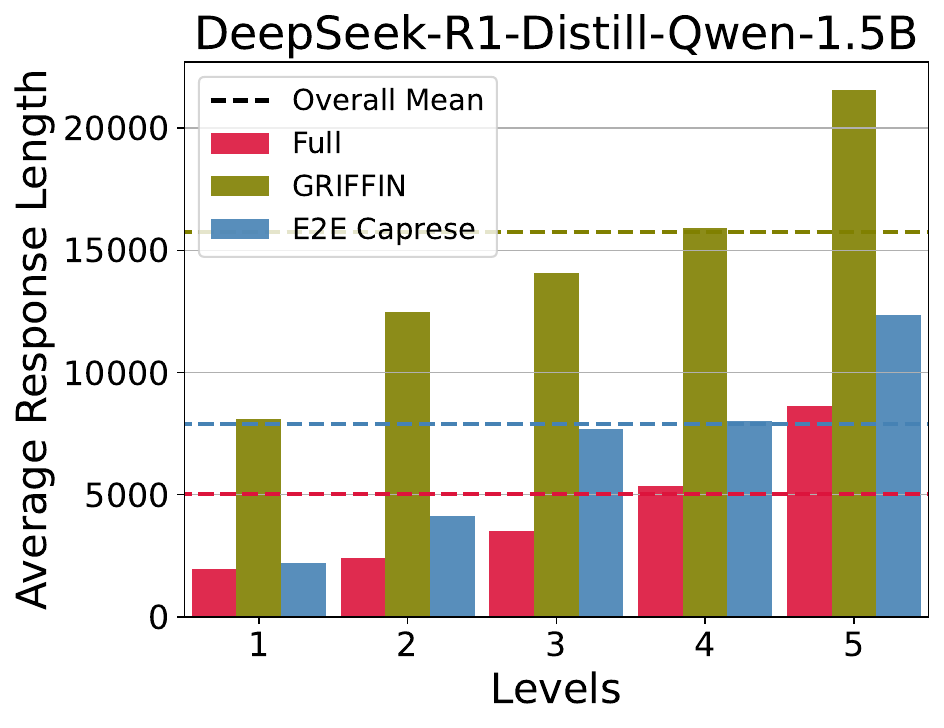}
\includegraphics[width=0.31\columnwidth]{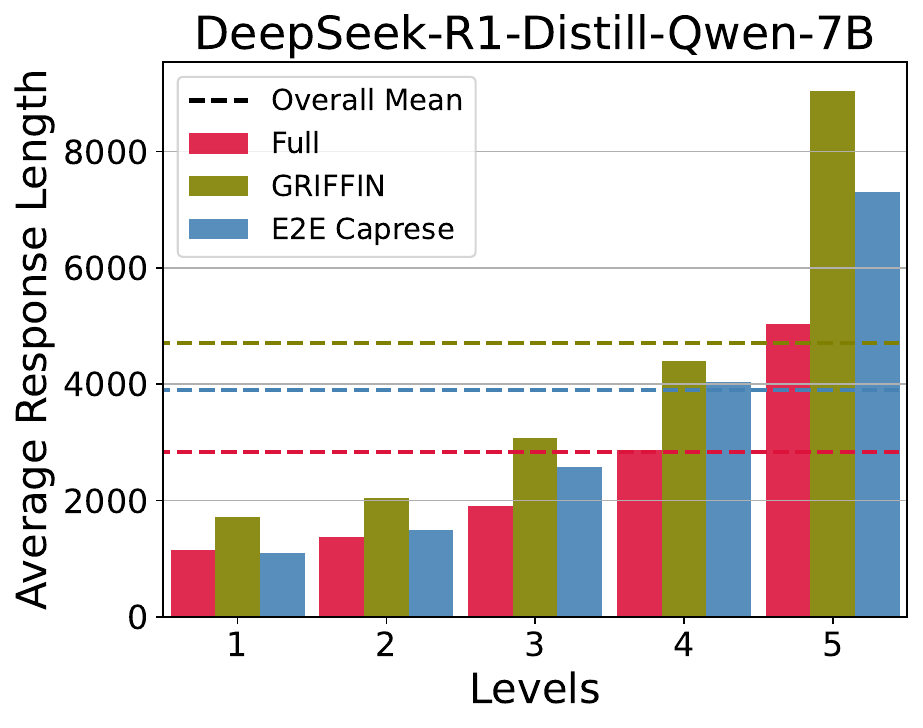}
\includegraphics[width=0.31\columnwidth]{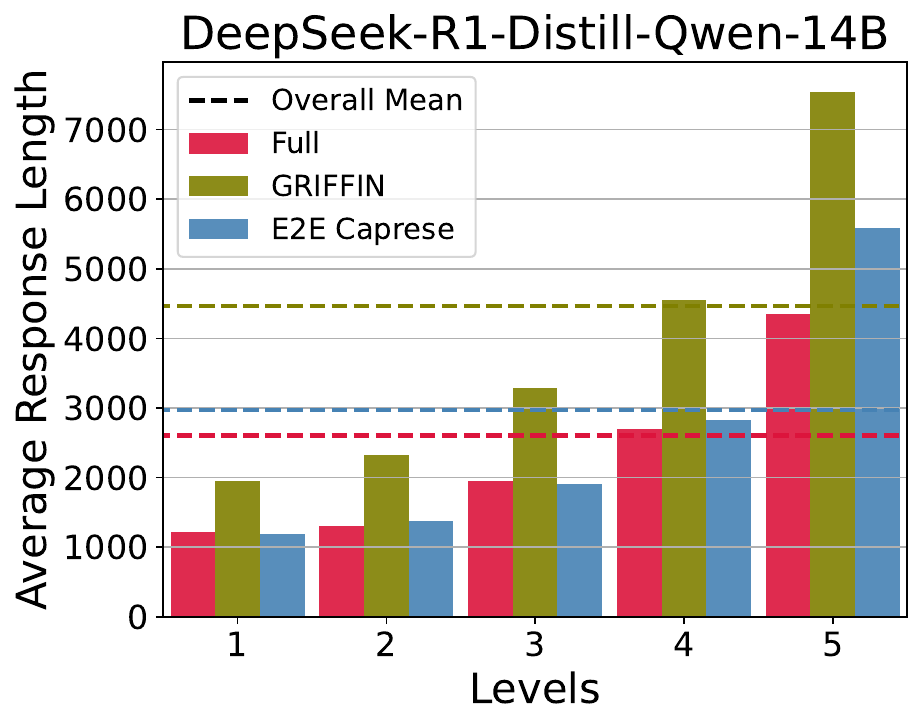}
\vspace{-0.12in}
\caption{Average number of response tokens for MATH-500 queries with increasing problem difficulty. The global averages are indicated by the dashed lines. Sparsity is set at 50\%.}
\label{fig:resp_lengths_griffin}
\end{center}
\vspace{-0.15in}
\end{figure}

Complementing Figure~\ref{fig:resp_lengths} for CATS, we show the natural response lengths to MATH-500 samples in Figure~\ref{fig:resp_lengths_griffin}. Here, we see the similar observations as in Section~\ref{sec:resp_length}, though the difference between \methodname{} (\griffin{}) and the full model is slightly larger but decreasing with model size. Even so, response lengths of \methodname{} is still significantly shorter than \griffin{}.

\section{Rank of Learned Parameters}

\begin{figure}[h]
\begin{center}
\includegraphics[width=0.34\columnwidth]{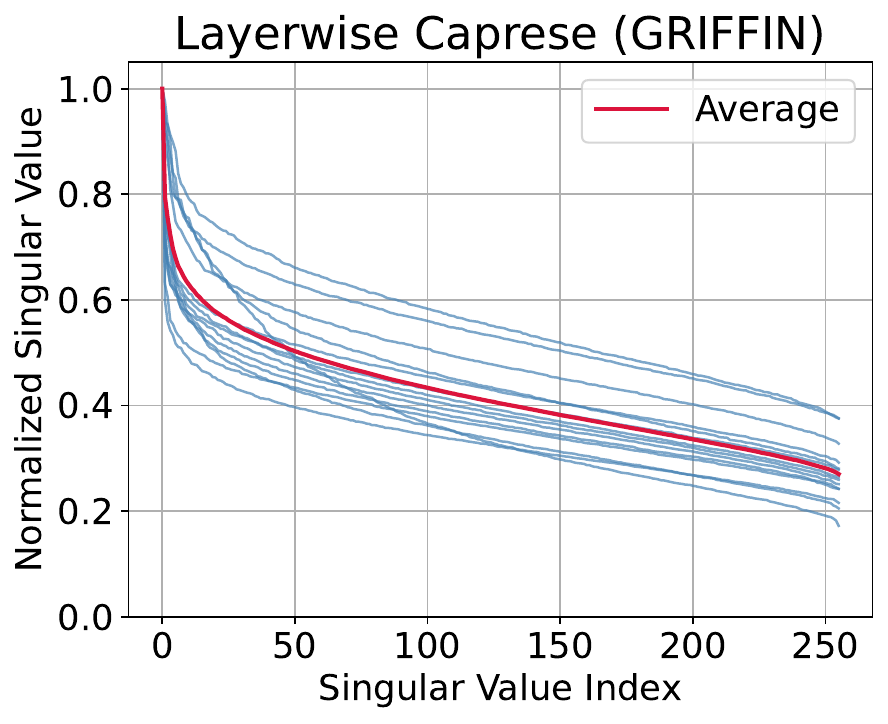}
\includegraphics[width=0.34\columnwidth]{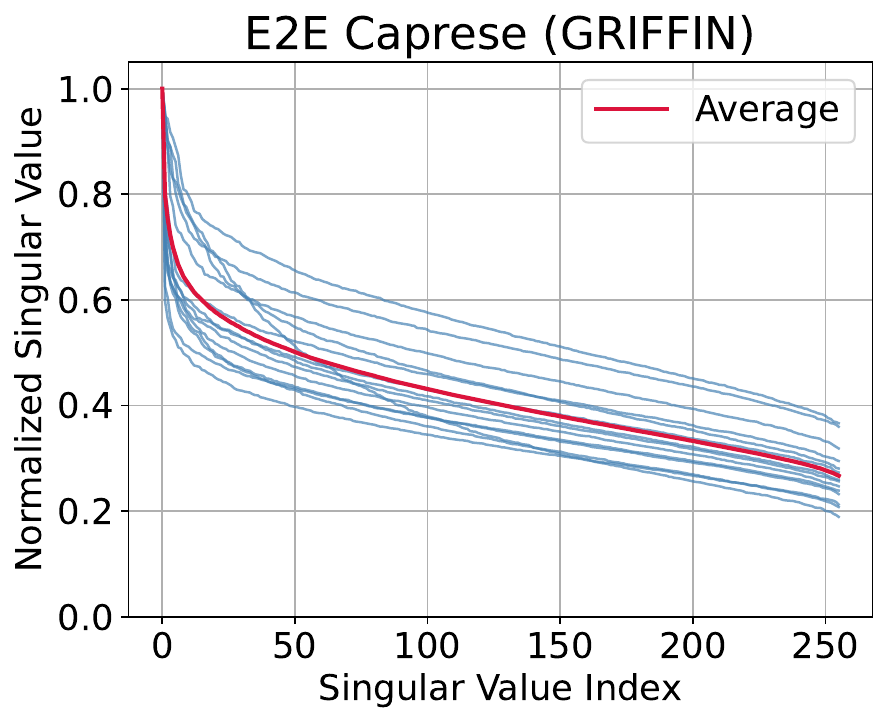}
\includegraphics[width=0.34\columnwidth]{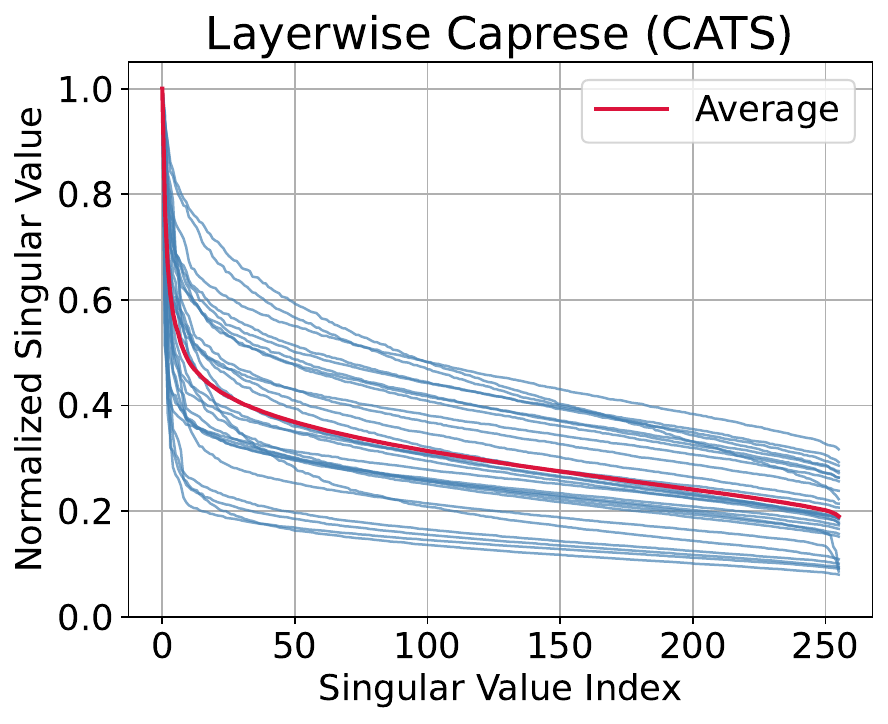}
\includegraphics[width=0.34\columnwidth]{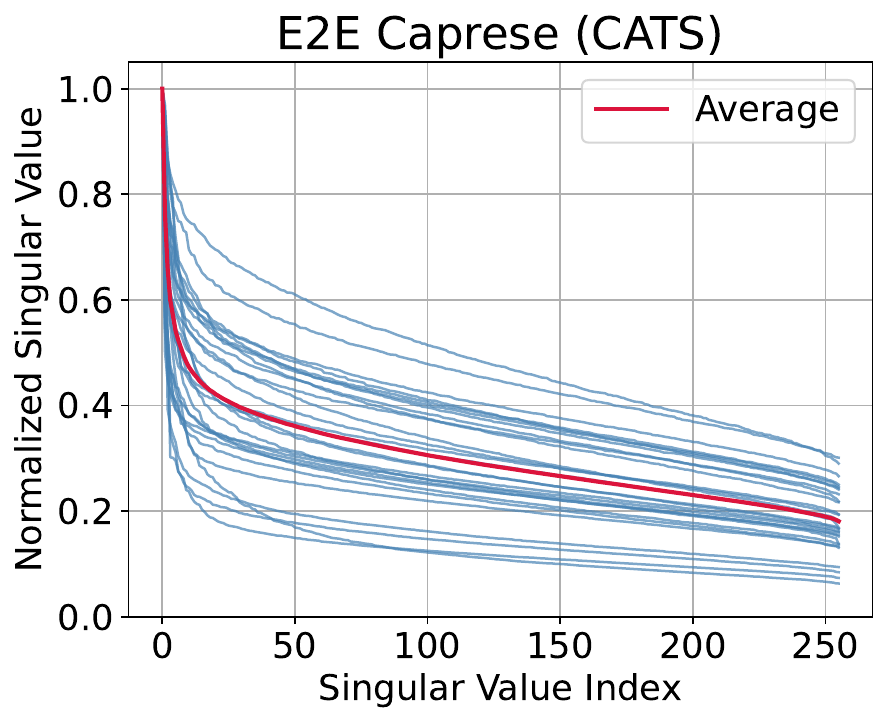}
\vspace{-0.12in}
\caption{Relative singular values of learned Llama 3.2 3B Instruct \methodname{} layers. Blue lines are individual layers; red lines are averaged across layers.}
\label{fig:param_rank}
\end{center}
\end{figure}

In Figure~\ref{fig:param_rank}, we plot the relative singular values for each learned product $\bL \bR$ from \eqref{eq:layerwise_obj} in Llama 3.2 3B Instruct. Although we set the inner rank to be 256, we see that some layers in \methodname{}(CATS) are still very low rank. In comparison, \methodname{}(\griffin{}) layers are relatively high rank, likely to due to the fact that \griffin{} performs highly structured FF sparsification. From this, we hypothesize that \griffin{} may better utilize an increased \methodname{} inner dimension.

\newpage
\section{Training Hyperparameters}
\label{app:hyperparameters}

Table~\ref{tab:hyperparams} lists the hyperparameter settings for training \methodname{} layers. The E2E learning rates lie in the interval [4e-6, 2e-4], where larger models tend to learn better with smaller learning rates.

\begin{table}[h]
\caption{\methodname{} layer-wise and E2E training hyperparameters.}
\begin{center}
\begin{tabular}{lcc}
\toprule
& Layer-wise & End-to-end \\
\midrule
Optimizer & Adam & Adam \\
Learning rate & 1e-3 & [4e-6, 2e-4] (varies) \\
Batch size & 128 & 16 \\
Epochs & 20 & 3 \\
Training samples & 2e5 & 2e5 \\
Scheduler & Linear & Linear \\
Warmup & 2\% & 2\% \\
\bottomrule
\end{tabular}
\end{center}
\label{tab:hyperparams}
\end{table}

\end{document}